\documentclass[11pt,a4paper]{article}

\usepackage{booktabs}
\usepackage{multirow}
\usepackage{subfigure}
\usepackage{multirow}
\usepackage{float}
\usepackage{wrapfig}
\usepackage[style = alphabetic,natbib=true,citestyle=alphabetic,maxbibnames=99,backend=biber,sorting=nyt,backref=true]{biblatex} %

\usepackage{filecontents}

\DefineBibliographyStrings{english}{%
 backrefpage = {Cited on page},
 backrefpages = {Cited on pages},
}
\usepackage[colorlinks,linkcolor = RawSienna, urlcolor  = Sepia, citecolor = RoyalBlue, anchorcolor = ForestGreen,bookmarks=False]{hyperref}
\usepackage[dvipsnames]{xcolor}

\usepackage{amsthm}
\usepackage{amsmath}
\usepackage{amssymb}
\usepackage{spencercommands}
\usepackage{fullpage,times,url,caption}

\usepackage[shortlabels]{enumitem}
\usepackage{enumerate}

\usepackage{bm}
\newcommand{\vw}{\bm{w}}
\newcommand{\vu}{\bm{u}}
\newcommand{\va}{\bm{a}}
\newcommand{\vx}{\boldsymbol{x}}
\newcommand{\mW}{\bm{W}}
\newcommand{\mU}{\bm{U}}
\newcommand{\mI}{\bm{I}}

\newcommand{\KL}{D_{\mathrm{KL}}}
\newcommand*{\dif}{\,\mathrm{d}}

\DeclareMathOperator*{\argmax}{arg\,max}

\newtheorem{theorem}{Theorem}[section]

\newtheorem{lemma}[theorem]{Lemma}

\newtheorem{definition}[theorem]{Definition}

\newtheorem{assumption}[theorem]{Assumption}

\author{
     Shen-Huan Lyu\thanks{Nanjing University. E-mail: {\tt lvsh@lamda.nju.edu.cn}}
     \and
     Lu Wang\thanks{Nanjing University. E-mail: {\tt wangl@lamda.nju.edu.cn}}
     \and
     Zhi-Hua Zhou\thanks{Corresponding author. Nanjing University. E-mail: {\tt zhouzh@lamda.nju.edu.cn}}
}
\date{}

\title{\textbf{Improving Generalization of Deep Neural Networks by Leveraging Margin Distribution}\thanks{This is a preprinted manuscript.}}
\allowdisplaybreaks[4]
\begin{document}

\begin{titlepage}

  \large
  \hfill
  \vfill
  \vspace*{2.0cm}
  \begin{center}
    \textcolor{Sepia}{\LARGE\textbf{Improving Generalization of Deep Neural Networks by Leveraging Margin Distribution}}
  \end{center}
  \vspace{0.5cm}
  \hrule
  \vspace{1.5cm}

  \begin{center}
    \begin{minipage}[t]{1.0\textwidth}
      \begin{flushleft}
        \emph{Author:}\\
        \href{http://www.lamda.nju.edu.cn/lvsh/}{{Shen-Huan} \textsc{Lyu}},
        \href{http://www.lamda.nju.edu.cn/wangl/}{{Lu} \textsc{Wang}}, and
        \href{http://www.lamda.nju.edu.cn/zhouzh/}{{Zhi-Hua} \textsc{Zhou}} (Corresponding author)
      \end{flushleft}
    \end{minipage}\\[2.5cm]
    
    \textbf{A pre-printed manuscript\\
    which is accepted by Neural Networks}\\[1.5cm]
    
    \includegraphics[width=5cm]{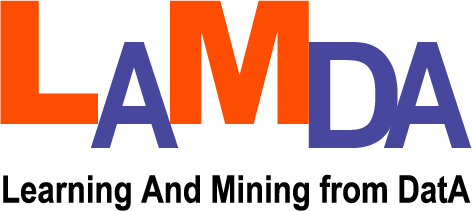}\\
    \href{http://www.lamda.nju.edu.cn/CH.MainPage.ashx}{LAMDA Group}\\
    Department of Computer Science and Technology\\
    National Key Laboratory for Novel Software Technology\\[0.5cm]

    \hfill
    \vfill

    \includegraphics[width=3cm]{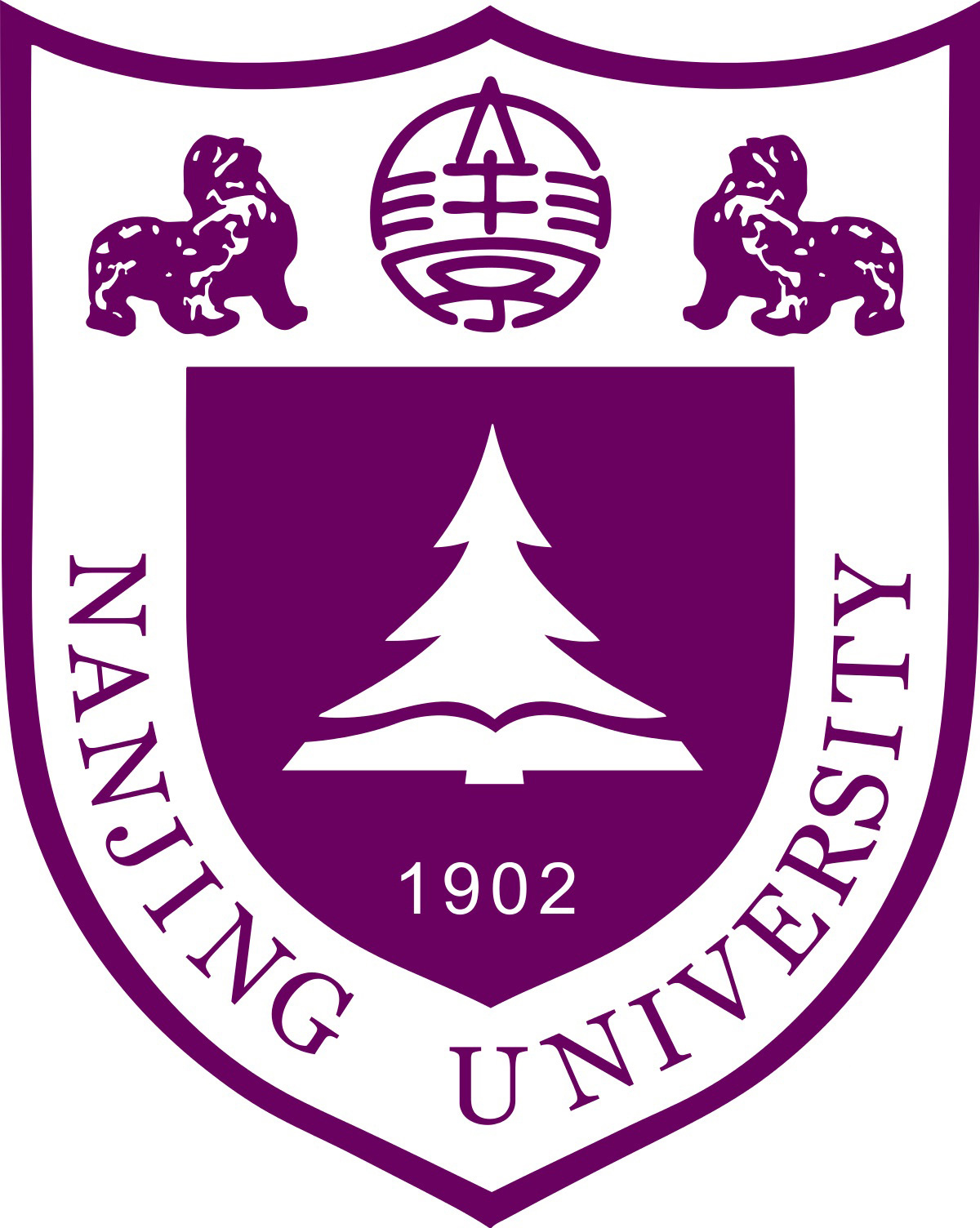}\\
    \href{https://www.nju.edu.cn/}{Nanjing University}\\
    Nanjing, China\\[2em]
    \begin{flushleft}
    {\footnotesize \textbf{Citation:} Shen-Huan Lyu, Lu Wang, and Zhi-Hua Zhou. Improving generalization of deep neural networks by leveraging margin distribution. Neural Networks, 2022. doi: https://doi.org/10.1016/j.neunet.2022.03.019.}
    \end{flushleft}
  \end{center}
\end{titlepage}

{\hypersetup{linkcolor=Black}\tableofcontents}
\thispagestyle{empty}
	
\maketitle
\setcounter{page}{1}

\begin{abstract}
Recent research has used margin theory to analyze the generalization performance for deep neural networks (DNNs). The existed results are almost based on the spectrally-normalized minimum margin. However, optimizing the minimum margin ignores a mass of information about the entire margin distribution, which is crucial to generalization performance. In this paper, we prove a generalization upper bound dominated by the statistics of the entire margin distribution. Compared with the minimum margin bounds, our bound highlights an important measure for controlling the complexity, which is the ratio of the margin standard deviation to the expected margin. We utilize a convex margin distribution loss function on the deep neural networks to validate our theoretical results by optimizing the margin ratio. Experiments and visualizations confirm the effectiveness of our approach and the correlation between generalization gap and margin ratio.
\end{abstract}

\section{Introduction}

Deep neural networks (DNNs) are making major advances in solving problems that have resisted the best attempts of the artificial intelligence community for many years \citep{lecun15deep}, especially in the field of computer vision \citep{gori2022ten}. Recently, many research try to explain the practical success of DNNs via generalization, which is the ability of a classifier to perform well on unseen samples. However, some new empirical evidence has started to question this explanation. Adversarial training samples can cause the model to misclassify seriously by slight feature perturbation \citep{goodfellow2015explaining,papernot2017practical}. On the other hand, \citet{zhang2021understanding} find that the deep neural networks have enough complexity to fit an arbitrarily corrupted data, and a small geometric transformation may cause networks deteriorating in performance \citep{azulay2019why}. This complex and fragile nature of DNNs leads to a key problem, how to use the \textit{data distribution} and \textit{network parameters} to estimate the generalization ability of DNNs. Although several regularization techniques, such as dropout \citep{srivastava14dropout}, batch normalization \citep{ioffe15batch}, and weight decay \citep{krogh92simple}, do improve the generalization performance of the over-parameterized deep models, \citet{zhang2021understanding} show that these regularizers cannot solve this problem either.

Consequently, several recent works \citep{neyshabur15norm,bartlett17spectrally,neyshabur18spectrally,arora18compression} have started to address this question, proving that we can control the capacity of DNNs via different upper bounds based on the minimum margin. However, the generalization bounds based on analyses of model complexity and noise stability only focus on the minimum margin, which is based on the closest distance of the training points to the decision boundary. This notion is brittle and sensitive to outliers due to a lack of the entire margin distribution information. 
\citet{jiang19predicting} propose a measure by looking at the entire distribution of distances, and conduct empirical studies on how well it can predict the generalization gap. However, how the margin distribution information affects the generalization error of the model still needs more specific theoretical analysis, which will lead us to optimize the entire margin distribution appropriately.

The margin distribution has been shown to correspond to generalization properties in the literature on linear models and boosting algorithms, \citet{schapire97boosting} first introduce it to explain the phenomenon that AdaBoost seems resistant to overfitting problem. 
Two years later, \citet{breiman99arc} indicates that the minimum margin is crucial for margin theory. \citet{reyzin06boosting} conjecture that the margin distribution, rather than the minimum margin, plays a key role. The debate has been finally solved by \citet{gao13boosting} who theoretically proved that the AdaBoost process attempts to maximize the margin mean and minimize margin variance simultaneously; highlighting for the first time that two factors rather than a single factor are crucial for margin theory. These two factors are the first and second-order statistics describing the margin distribution, while in most cases the higher-order one is less useful. Their result successfully explains why AdaBoost seems resistant to overfitting: even when the training error reaches zero, the margin mean can be increased and/or the margin variance can be decreased further, leading to the improvement of generalization performance; it also discloses that AdaBoost will finally overfit: when the margin mean cannot be increased and margin variance cannot be decreased further. The long march of the theoretical exploration of AdaBoost is summarized in \citet{zhou2014large}, and \citet{gao13boosting}’s result has been theoretically confirmed by \citet{gronlund2019margin}. Inspired by \citet{gao13boosting}’s finding, powerful learning machines can be built by maximizing the margin mean and minimizing margin variance simultaneously, rather than simply maximizing the minimum margin like in traditional large-margin machines.
\citet{zhang16odm} propose the optimal margin distribution machine (ODM) for binary classification. In \citet{zhang17mcodm,zhang18optimal,zhang18semi,tan2020multi,zhang2020optimal}, ODM is extended to many forms.

In this paper, we study a $d$-layer feed-forward network with ReLU activation functions. Our theoretical result states that the statistics of margin distribution play an important role in the generalization estimation rather than the traditional minimum margin. This result is consistent with the previous results on boosting and linear algorithms \citep{schapire97boosting,gao13boosting,zhang16odm}. It also inspires us to understand the similarities between deep learning and traditional machine learning from the perspective of margin distribution. Specially, we propose a new loss function to optimize the statistics-based measure in the theoretical results. A strong correlation between generalization and our measure is empirically shown by studying a wide range of network structures trained on the MNIST, CIFAR-10 and ImageNet datasets.
The detailed contributions of this paper are as follows:

\paragraph{\textbf{\emph{PAC guarantee}}} Our bound shows that we can restrict the capacity of deep nets by the ratio of second- to first-order statistic of margin distribution at the last layer. Compared with the existing results based on minimum margin \citep{bartlett17spectrally,neyshabur18spectrally,arora18compression}, our bound contains more information on the entire margin distribution to estimate the generalization error. Moreover, the empirical evaluation shows that optimizing the margin ratio can control the model capacity to alleviate the overfitting risk. 

\paragraph{\textbf{\emph{Optimization}}} Inspired by our theoretical result, we encourage DNNs to optimize the margin ratio for better generalization performance. Therefore, we propose a new approach called \underline{m}argin \underline{d}istribution \underline{Net}works (mdNet), which utilizes a convex margin distribution loss function to optimize the first- and second-order statistics of margin. Moreover, we empirically evaluate our loss function on deep neural networks across different image datasets and model structures. Specifically, empirical results demonstrate the effectiveness of the proposed method in learning tasks with limited training data.\\

The rest of paper is organized as follows. The related work is introduced in Section~\ref{sec:re}. Some notations are introduced in Section~\ref{sec:no}. In
Section~\ref{sec:md}, we present a generalization bound leveraging margin distribution rather than minimum margin and demonstrate that the ratio of the margin standard deviation to the expected margin is the key to control the model capacity. Section~\ref{sec:pr} list the detailed proofs for our theorems and lemmas. In Section~\ref{method}, we formulate the convex loss function to optimize the margin ratio. Section~\ref{experiment} reports our experimental studies and empirical observations.  Finally, Section~\ref{sec:cc} concludes with future work.

\section{Related work} \label{sec:re}

Recently, margin-based deep learning algorithms have developed rapidly. \citet{schroff15facenet} use the triplet loss to encourage a distance constraint similar to the contrastive loss. Similarly, \citet{chan15pcanet} enhance the supervision of the learned filters by incorporating the information of class labels in the training data and learn the filters based on the idea of multi-class linear discriminant analysis (LDA) for classification task. \citet{liu16large} propose a generalized large-margin softmax loss which explicitly encourages \textit{intra-class compactness} and \textit{inter-class separability} in the learned representation space. It would be interesting to theoretically study feature space transformation which might be a key to understanding mysteries behind the successes of deep neural networks \citep{zhou2021over}.
Since \citet{bartlett17spectrally} and \citet{arora18compression} associate the generalization of deep neural networks with the minimum margin, a line of work establishes that first-order methods can automatically maximizing the minimum margin in the settings of logistic regression \citep{gunasekar2018characterizing}, deep linear networks \citep{soudry2018implicit,gunasekar2018implicit,ji2019gradient,li2018algorithmic}, and symmetric matrix factorization \citep{li2018algorithmic}. However, \citet{wei2018margin} point that how to extend these results to non-linear neural networks remains unclear. Recently, \citet{wu2021learning} propose to understand the model dynamics from the perspective of control theory. Another line of algorithm-dependent analysis of generalization \citep{HardtRS16,Mou0Z018,chen2018stability} uses stability of specific optimization algorithms that satisfy certain generic properties
like convexity, smoothness, etc. Specially, \citet{KeskarMNST17,DinhPBB17,ZhuWYWM19} make a connection between the sharpness of the solution obtained using the SGD algorithm and its ability to generalize well. The notion of sharpness corresponds to robustness to adversarial perturbations of parameters. Furthermore, \citet{NeyshaburBMS17,neyshabur18spectrally} draw a connection to the PAC-Bayesian theory for sharpness. The margin distribution measure presented in this paper is closely related to sharpness \citep{KeskarMNST17}, because we use the statistics of the margin distribution to theoretically describe the value of the allowable perturbation. Compared with the sharpness measure which is difficult to optimize, the margin distribution measure proposed in this paper is easy to calculate, and can be directly optimized through the SGD algorithm by designing a convex loss function. Recently, \citet{jiang19predicting} present abundant empirical evidence to validate that the generalization in deep learning can be estimated from the margin statistics. In addition, the relevant theories of domain adaptation \citep{mansour2009domain,zhang2012generalization,mansour2014robust} are also used to improve the generalization capability of deep learning \citep{pan2009domain,becker2013nonlinear,koniusz2017domain,rozantsev2019beyond}. Domain generalization cannot see existing training source domains during training. This makes domain generalization more challenging than domain adaptation but more realistic and favorable in practical applications \citep{ghifary2015domain,dubey2021adaptive,wang2021generalization,matskevych2022from}.

\section{Notations} \label{sec:no}

Consider the multi-class task with feature domain $\mathcal{X}$ and label domain $\mathcal{Y}$. Let $\mathcal{D}$ be an unknown (underlying) distribution over $\mathcal{X}\times\mathcal{Y}$. A training set $S=\{(\vx_1,y_1),\dots ,(\vx_m,y_m)\}$ and a validation set $S'=\{(\vx_1,y_1),\dots,(\vx_{m'},y_{m'})\}$ are drawn identically and independently  according to $\mathcal{D}$. We denote a labeled sample as $(\vx,y)\in\mathcal{D}$. 

Let $f_{\vw} \colon \mathcal{X}_{B,n} \rightarrow \mathcal{Y}'$ be the function represented by a $d$-layer feed-forward network with parameters \[\vw=\{\mW_1,\mW_2,\dots,\mW_d\}\] and output domain $\mathcal{Y}'=\mathbb{R}^k$. The entire network can be formulated as \[f_{\vw}(\vx)=\mW_d\phi(\mW_{d-1}\phi(\dots\phi(\mW_1\vx))),\] where $\phi$ is the ReLU activation function and let $\rho$ be an upper bound on the number of output units in each layer. 

We can define the fully connected networks (FNNs) recursively: \[\vx^1=\mW_1\vx \quad\text{and}\quad \vx^i=\mW_i\phi(\vx^{i-1}),\] where $\vx^i$ denotes the output of the $i$-th layer. 

The predicted label is denoted by 
\[h(\vx)=\argmax_j f_{\vw,j}(\vx)\in \mathcal{H},\]
where $h  \colon \mathcal{X} \rightarrow \mathcal{Y}$ is a map from the feature domain to the label domain and $f_{\vw,j}$ is the $j$-th element of the score vector. 

In the multi-class setting \citep[Chapter 9.2]{mohri2018foundations}, the label associated to point $\vx$ is the one resulting in the largest score $h(\vx)=\arg\max_{i}f_{\vw,i}(\vx)$. This naturally leads to the following definition of the margin $\gamma_h(\vx,y)$ of the function $h$ at a labeled example $(\vx,y)$: 
\begin{equation}\label{eq:margin}
	\gamma_{h}(\boldsymbol{x}, y)=f_{\vw,y}(\boldsymbol{x})-\max _{j \neq y} f_{\vw,j}(\boldsymbol{x}).
\end{equation}
Thus, $h$ misclassifies $(\vx,y)$ iff $\gamma_h(\vx,y)<0$.

\section{Margin distribution rather than minimum margin} \label{sec:md}

In Subsection~\ref{4.1}, we list error-resilience assumptions that will be used. In Subsection~\ref{4.2}, we introduce the existed results based on the minimum margin. In Subsection~\ref{4.3}, we present our main results based on the entire margin distribution. 

\subsection{Error-resilience assumptions}\label{4.1}

Here we formalize the error-resilience properties for deep neural networks. \citet{arora18compression} show that if we inject a scaled Gaussian noise to the input of deep nets, as it propagates up, the noise has rapidly decreasing effect on higher layers. This fact implies \emph{compressibility} of deep nets, i.e., low rank of parameters' matrix. The empirical version of noise-sensitivity parameters are first proposed by \citet{arora18compression}. It inspires us to bound the perturbation caused by Gaussian noise with the validation-based version of noise-sensitivity parameters below. 

\begin{assumption}\label{ass:1}
	(Layer Cushion). The layer cushion of layer $i$ is defined to be largest number $\mu_i$ such that for any validation data $\vx \in S'$:
	\begin{equation}
		\label{eq3}\mu_i\|\mW_i\|_F\|\phi(\vx^{i-1})\|_2 \leq \|\vx^{i}\|_2.
	\end{equation}
\end{assumption}

\begin{assumption}\label{ass:2}
	(Interlayer Cushion). For any two layers $i<j$, we define the interlayer cushion $\mu_{i,j}$, as the largest number such that for any validation data $\vx \in S'$:
	\begin{equation}
		\label{eq4} \mu_{i,j}\|J_{\vx^{i}}^{i,j}\|_F\|\phi(\vx^{i-1})\|_2  \leq \|\vx^{j}\|_2.
	\end{equation}
\end{assumption}

Furthermore, for any layer $i$ we define the minimal interlayer cushion as $\mu_{i \rightarrow} = \min_{i\leq j\leq L}\mu_{i,j} = \min\{ \frac{1}{\sqrt{\rho}}, \min_{i\leq j\leq L}\mu_{i,j} \}$. For any two layer $i<j$, denote by $M^{i,j}$ the operator for composition of these layers and $J_{\vx}^{i,j}$ be the Jacobian matrix (the partial derivative) of this operator at input $\vx$. Therefore, we have $\vx^{j}=M^{i,j}(\vx^{i})$. Furthermore, since the activation functions are ReLU (hence piece-wise linear), we have $M^{i,j}(\vx^{i})=J_{\vx^{i}}^{i,j}{\vx^{i}}$.

\begin{assumption}\label{ass:6}
	(Interlayer Smoothness). For any two layers $i<j$, we define the interlayer smoothness $\rho_{\delta}$ as the smallest number such that with probability $1-\delta$ over noise $\eta$ for any validation data $\vx\in S'$:
	\begin{equation}
		\|M^{i, j}(\vx^{i}+\eta)-J_{\vx^{i}}^{i, j}(\vx^{i}+\eta)\| \leq \frac{\|\eta\|\|\vx^{j}\|}{\rho_{\delta}\|\vx^{i}\|}
	\end{equation}
\end{assumption}
For a single layer, $\rho_{\delta}$ captures the ratio of input/weight alignment to noise/weight alignment. \citet{arora18compression} show that the interlayer smoothness is indeed good: $1/\rho_{\delta}$ is a small constant.

The next two conditions qualify a common appearance: if the input in the activation and margin calculations is well-distributed and the calculations do not correlate with the magnitude of the input, then one would expect that, the effect of applying activation at any layer and margin at last layer is to decrease the norm of the vector by at most some small constant factor, i.e., $c$ and $\alpha$.

\begin{assumption}\label{ass:3}
	(Activation Contraction). The activation contraction $c$ is defined as the smallest number such that for any layer $i$ and any validation data $\vx \in S'$:
	\begin{equation}
		\label{eq5} c\|\phi(\vx^{i})\|_2 \geq \|\vx^{i}\|_2.
	\end{equation}
\end{assumption}

\begin{assumption}\label{ass:4}
	(Margin Contraction). The margin contraction $\alpha$ is defined as the smallest number such that for any validation data $\vx \in S'$:
	\begin{equation}
		\alpha\|\gamma_{h}(\vx,y)\|_2 \geq \|\vx^d\|_2.
	\end{equation}
\end{assumption}

In this paper, we only use the noise-sensitivity parameters in Assumptions~\ref{ass:1} - \ref{ass:4} as descriptions of error-resilience properties, from which the margin distribution term of our bound is derived. Therefore, we just need estimate these parameters based on validation data to show the magnitude of our bound rather than optimizing these parameters in the training process like \citet{arora18compression} did. 

\subsection{Existed results}\label{4.2}

In the deep learning theory community, great efforts have been made to explain why over-parameterized deep neural networks can success, which is contrary to the classical VC dimension analysis \citep{bartlett1998almost,harvey17nearly}. \citet{bartlett17spectrally} and \citet{neyshabur18spectrally} made an important stride by showing minimum margin based bounds for multi-layer neural networks. These bounds do not depend directly on the number of parameters of the network but depends on the normalized minimum margin. Theorem \ref{theorem:spec} provides a unified description of these bound. The only difference between them lies in the value of constants and the type of norms.

\begin{theorem}\label{theorem:spec}\citep{bartlett17spectrally,neyshabur18spectrally}
	For any $d,\rho>0$ and $\|\vx\|_2\leq B$, let $f_{\vw}:\mathcal{X} \rightarrow \mathbb{R}^k$ be a $d$-layer feed-forward network with ReLU activation. Then, for any $\delta > 0$, with probability $\geq 1-\delta$ over a training set of size $m$, for any $\vw=\{\mW_1,\mW_2,\dots,\mW_d\}$, we have:
	\begin{equation}
		L_{0}\left(f_{\vw}\right) \leq \widehat{L}_{\gamma}\left(f_{\vw}\right)+\mathcal{O}\left(\sqrt{\frac{B^{2} d^{2} \rho \ln (d h) \Pi_{i=1}^{d}\left\|\mW_{i}\right\|_{2}^{2} \sum_{i=1}^{d} \frac{\left\|\mW_{i}\right\|_{F}^{2}}{\left\|\mW_{i}\right\|_{2}^{2}}+\ln \frac{d m}{\delta}}{\gamma^{2} m}}\right),
	\end{equation}
	where $L_0(\cdot)$ is the 0-1 loss, $\widehat{L}_{\gamma}(\cdot)=\Pr_{S} \left[\gamma_{h}(\vx,y) \leq \gamma   \right]$ is the empirical estimation of $\gamma$-margin loss and $\mathcal{O}(\cdot)$ describes the limiting behavior of a function.
\end{theorem}

Based on this margin theory view, \citet{arora18compression} provide an improved bound by considering the compressibility of deep nets as follows:

\begin{theorem}\label{theorem:compress}\citep{arora18compression}
	For any $d>0$, let $f_{\vw}:\mathcal{X} \rightarrow \mathbb{R}^k$ be a $d$-layer feed-forward network with ReLU activation. Then, for any $\delta > 0$, with probability $\geq 1-\delta$ over a training set of size m, for any $\vw=\{\mW_1,\mW_2,\dots,\mW_d\}$, we have:
	\begin{equation}
		L_{0}\left(f_{\vw}\right) \leq \widehat{L}_{\gamma}\left(f_{\vw}\right)+\mathcal{O}\left(\sqrt{\frac{c^{2} d^{2} \max _{\vx \in S}\left\|f_{\vw}(\vx)\right\|_{2}^{2} \sum_{i=1}^{d} \frac{1}{\mu_{i}^{2} \mu_{i\rightarrow}^{2}}}{\gamma^{2} m}}\right)
	\end{equation}
	where $\mu_i,\mu_{i \rightarrow},c,\alpha$ are layer cushion, interlayer cushion, activation contraction and interlayer smoothness defined in Assumptions~\ref{ass:1}, \ref{ass:2}, \ref{ass:3} and \ref{ass:4} respectively
\end{theorem}

These existed results follow the traditional margin theory, so they only focus on the minimum margin $\gamma$. Because they lack the description of the entire margin distribution, they can only take the minimum margin $\gamma$ as the optimization target to improve the generalization performance. These methods ignore the information of the entire margin distribution. In the next subsection, we expect to prove a bound related to the entire margin distribution, so as to inspire us to directly optimize margin distribution for improving the generalization performance for DNNs.

\subsection{Main results}\label{4.3}

\begin{figure}[h]
	\vskip 0.1in
	\centering
	\subfigure[The minimum margin based classifier (red line) and the margin distribution based classifier (black line). The blue triangle and green square represent the two classes of instances, while the dotted ellipses represent their underlying distribution. \label{fig:margin_distribution}]{
			\begin{minipage}[t]{0.31\columnwidth}
					\includegraphics[width=\columnwidth]{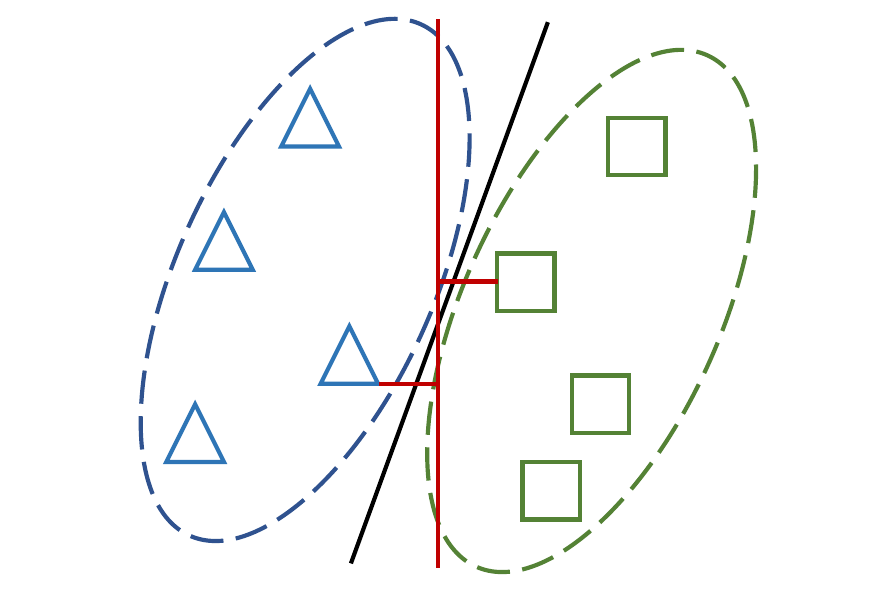}
				\end{minipage}
		}
	\hfill
	\subfigure[The $(r,\theta)$-margin distribution loss function (red line). The green dotted lines represent the confidence area of margins, which has zero loss. Lemma~\ref{Lemma 2} shows that the perturbation caused by $\vu$ (blue arrow) is related to the generalization error.\label{fig:margin_loss}]{
		\begin{minipage}[t]{0.33\columnwidth}
			\includegraphics[width=\columnwidth]{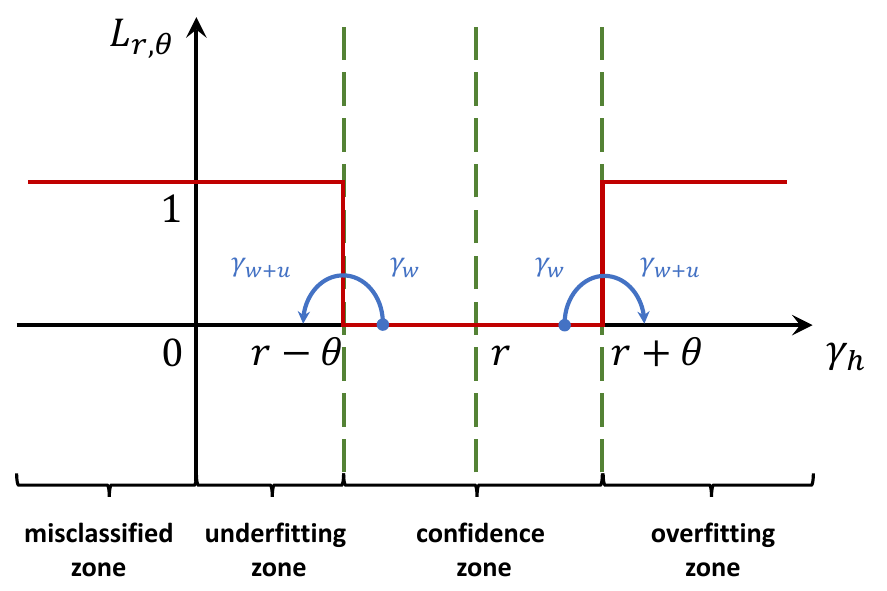}
		\end{minipage}
	}
    \hfill
	\subfigure[The convex margin distribution loss function (red line). This convex function is used as an alternative function of $(r,\theta)$-margin distribution loss function, so that the deep neural networks can optimize the margin distribution through SGD algorithm. \label{fig:convex_loss}]{
		\begin{minipage}[t]{0.31\columnwidth}
			\includegraphics[width=\columnwidth]{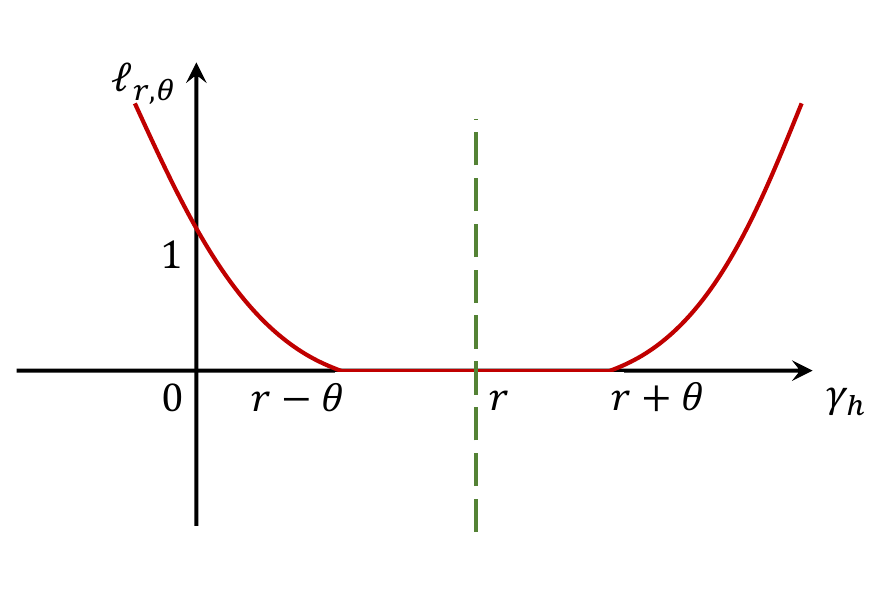}
		\end{minipage}
	}
		\caption{Illustration of the margin distribution analysis and loss functions.} 
		\label{fig:margin_view}
\end{figure}

We begin with an intuitive comparison of the minimum margin based classifier and the margin distribution based classifier. Figure~\ref{fig:margin_distribution} shows that maximizing the minimum margin will make the classifier easy to be misled by a small number of samples, thus ignoring the distribution information of samples, while the margin distribution based classifier considers the mean and variance of samples and generalizes better. 

In order to utilize the mean and variance information into the theoretical analysis, we design a new margin loss, which uses $r$ to adjust the mean of margin and $\theta$ to adjust the variance of margin. For any parameter $r>\theta>0$, we can define a $(r,\theta)$-margin distribution loss function (see Figure~\ref{fig:margin_loss}), which penalizes $h$ with a cost of $1$ when it predicts $\vx$ with a margin smaller than $r-\theta$, but also penalizes $h$ when it predicts $\vx$ with a margin larger than $r+\theta$. The margin distribution bound is presented in terms of this loss function, which is formally defined as follows.
\begin{definition}(Expected margin distribution loss function). 
	\label{def:5} For any $r>\theta>0$, the $(r,\theta)$-margin loss is the function $L_{r,\theta}(\cdot)$ defined for all $h\in \mathcal{H}$ as:
	\begin{equation}
		L_{r,\theta}(h) = \Pr_{\mathcal{D}} \left[\gamma_{h}(\vx,y) \leq r - \theta   \right] + \Pr_{\mathcal{D}} \left[ \gamma_{h}(\vx,y)   > r + \theta   \right] .
	\end{equation}
\end{definition}
Intuitively, our $(r,\theta)$-margin distribution loss function looks for a classifier $h$ which forces as many data points as possible into the \emph{zero-loss band} ($r-\theta\leq \gamma_{h}(\vx,y)< r+\theta$). Therefore, we let $r\simeq\mathbb{E}_{\mathcal{D}} [\gamma_{h}(\vx,y)], \theta^2\simeq\Var_{\mathcal{D}}[\gamma_{h}(\vx,y)]$, which implies that the expected margin is larger than the standard deviation. Actually, $\theta$ just need to be a second-order statistic, so we can re-scale $\theta^2=a\cdot\Var_{\mathcal{D}}[\gamma_{h}(\vx,y)]$ to satisfy $r>\theta$. In this way, the ($r,\theta$)-margin distribution loss is a surrogate loss function.
In particular, for $r = \theta$ and $\theta \rightarrow \infty$, the zero-loss band is the positive area ($\gamma_{h}(\vx,y)>0$) and $L_{r,\theta}$ corresponds to the 0-1 loss $L_0$. Let $\widehat{L}_{r,\theta}(f_{\vw})$ be the empirical estimate of the expected margin distribution loss. So we also denote the expected risk and the empirical risk as $L_0(f_{\vw})$ and $\widehat{L}_0(f_{\vw})$, which are bounded between 0 and 1. 

We begin with bounding the change of output caused by the noise on the classifier $\vu$ with the noise-sensitivity parameters and the statistics of margin distribution:

\begin{lemma}\label{Lemma 1}
	Let $f_{\vw}\colon\mathcal{X}\rightarrow\mathbb{R}^k$ be a $d$-layer network. For any $d > 0$,
	and $vec(\{\mU_i\}_{i=1}^{d})=(\mU_1,\mU_2,\dots,\mU_d)$ is a vector of perturbation parameters with $\mU_i=\bm{\beta}_i\|\mW_i\|_F,$ and $\bm{\beta}=vec(\{\bm{\beta}_i\}_{i=1}^{d})=(\bm{\beta}_1,\bm{\beta}_2,\dots,\bm{\beta}_d)$ is a vector of random vectors with $\mathbb{E}[\bm{\beta}\bm{\beta}^\top]=\sigma^2\mI$, the change of the output of the network can be bounded with a fixed probability ($\delta=1/2$):
	\begin{equation}
		\max_{\vx\in \mathcal{X}}|f_{\vw+\vu}(\vx)-f_{\vw}(\vx)|_2^2 \leq \mathcal{O}\left(
		\sum_{i=1}^{d}\frac{d\alpha^2c^2\sigma^2(r+\theta)^2}{\mu_{i}^2\mu_{i \rightarrow}^2}\right).
	\end{equation}
\end{lemma}

\begin{figure}[h]
	\vskip 0.0in
	\centering
	\subfigure[$\sigma^2\propto\left(\frac{1-\lambda}{1+\lambda}\right)^2$.]{
		\begin{minipage}[t]{0.25\columnwidth}
			\includegraphics[width=\columnwidth]{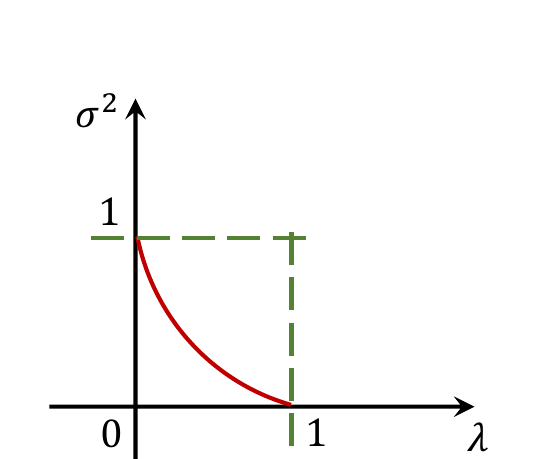}
		\end{minipage}
	}
	\hfill
	\subfigure[Fixed $r$, larger $\theta$, larger $\lambda$, smaller $\sigma^2$.]{
		\begin{minipage}[t]{0.27\columnwidth}
			\includegraphics[width=\columnwidth]{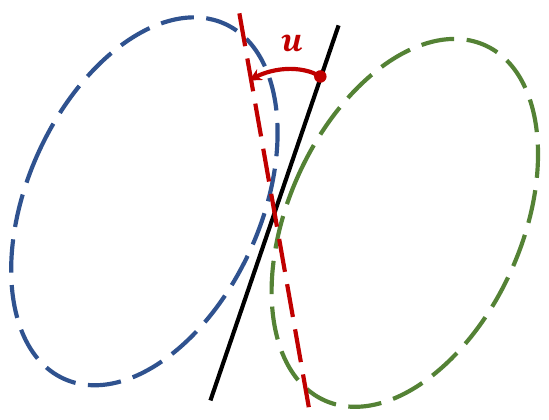}
		\end{minipage}
	}
	\hfill
	\subfigure[Fixed $r$, smaller $\theta$, smaller $\lambda$, larger $\sigma^2$.]{
		\begin{minipage}[t]{0.28\columnwidth}
			\includegraphics[width=\columnwidth]{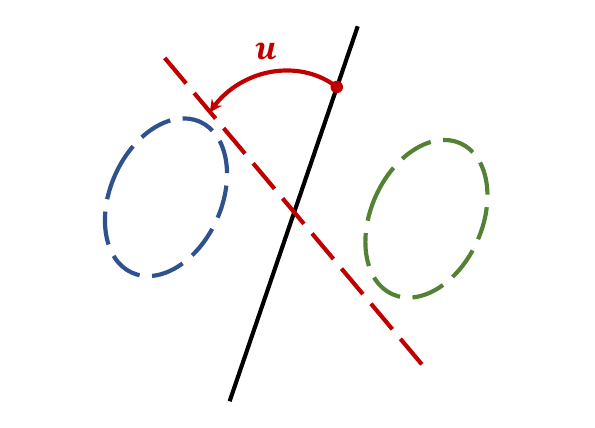}
		\end{minipage}
	}
	\caption{Illustration of the relationship between margin distribution and allowable perturbation.} 
	\label{fig:lambda}
\end{figure}

The result shows that the perturbation caused by $\vu$ increases with the variance $\sigma^2$ is related to the outermost edge of margin distribution $r+\theta$, that is, the right green dotted line in Figure~\ref{fig:margin_loss}. The next Lemma shows that we can bound the generalization gap through a Kullback-Leibler divergence term, if we can guarantee that the perturbation caused by $\vu$ is smaller than $\frac{r-\theta}{8}$ with a constant probability. Therefore, the allowable value of $\sigma^2$ under the $(r,\theta)$-margin distribution assumption is consistent with the intuitive understanding (see Figure~\ref{fig:lambda}), i.e., $\sigma^2\propto\frac{r-\theta}{r+\theta}=\frac{1-\lambda}{1+\lambda}$, where $\lambda=\theta/r\in(0,1)$. When the margin distribution is more compact (smaller $\lambda$), the larger noise $\sigma^2$ can be allowed, that is, it is not easy to cause misclassification. When the margin distribution is more loose (larger $\lambda$), even a small noise have misclassification risk.

\begin{lemma}\label{Lemma 2}
	Let $f_{\vw}:\mathcal{X} \rightarrow \mathbb{R}^k$ be any predictor with parameters $\vw$, and $P$ be any distribution on the parameters that is independent of the training data. Then, for any $r > \theta > 0, \ \delta > 0$, with probability at least $ 1-\delta$ over the training set of size $m$, for any $\vw$, and any random perturbation $\vu$ \mbox{s.t.} $\Pr_{\vu} \left[ \max_{\vx\in \mathcal{X}}| f_{\vw+\vu}(\vx) - f_{\vw}(\vx) |_{2} < \frac{r-\theta}{8} \right] \geq \frac{1}{2}$, we have:
	\begin{equation}
		L_0(f_{\vw}) \leq \widehat{L}_{r,\theta}(f_{\vw}) +  4\sqrt{\frac{\KL(\vw+\vu\Vert P) + \ln \frac{6m}{\delta}}{m-1}}.
	\end{equation}
\end{lemma}

The detailed proof is presented in Section~\ref{sec:4.2}. This Lemma improves the result of \citet[Lemma 1]{neyshabur18spectrally}, especially using two parameters $\theta, r$ to describe the entire margin distribution instead of using one parameter $\gamma$ to describe the minimum margin. Based on this result, we can derive the following generalization bound, with proof deferred to Section~\ref{sec:4.3} showing that the Kullback-Leibler divergence term is inversely proportional to $\sigma^2$.

\begin{theorem}\label{theorem 1}
	For any $d,\rho>0$, let $f_{\vw}:\mathcal{X} \rightarrow \mathbb{R}^k$ be a $d$-layer feed-forward network with ReLU activation. Then, for any $\delta > 0$, with probability $\geq 1-\delta$ over a training set of size m, for any $\vw$, we have:
	\begin{equation}
		L_0(f_{\vw}) \leq \widehat{L}_{r,\theta}(f_{\vw}) +
		\mathcal{O} \left( \sqrt{\frac{\left(\frac{1+\lambda}{1-\lambda}\right)^2\left(\sum_{i=1}^{d} \frac{d\alpha^2c^2}{\mu_{i}^2\mu_{i \rightarrow}^2}\right)+\ln\frac{6m}{\delta}}{m}} \right).
	\end{equation}
	where the margin ratio is defined by $\lambda=\theta/r.$
\end{theorem}

We prove an upper bound on generalization gap related to the margin ratio term, where $\lambda$ is a parameter denoting the ratio of the margin standard deviation $\theta$ to the expected margin $r$ over the underlying distribution, and the error-resilience term relies on the noise sensitivity \citep{arora18compression} quantified by $\mu_i,\mu_{i \rightarrow},c,\alpha$ (See Assumptions~\ref{ass:1}, \ref{ass:2}, \ref{ass:3} and \ref{ass:4}). 
Theorem~\ref{theorem 1} states that the entire margin distribution has much leverage in generalization performance rather than the minimum margin.
Specifically, restricting a smaller $\lambda$ (larger $r$ and smaller $\theta$) can effectively control the capacity of models, so as to reduce the risk of overfitting. 
It inspires us that optimizing margin distribution can get better generalization performance than the traditional minimum margin maximization algorithm.


\begin{figure}[!ht]
    \centering
    \includegraphics[width=0.5\textwidth]{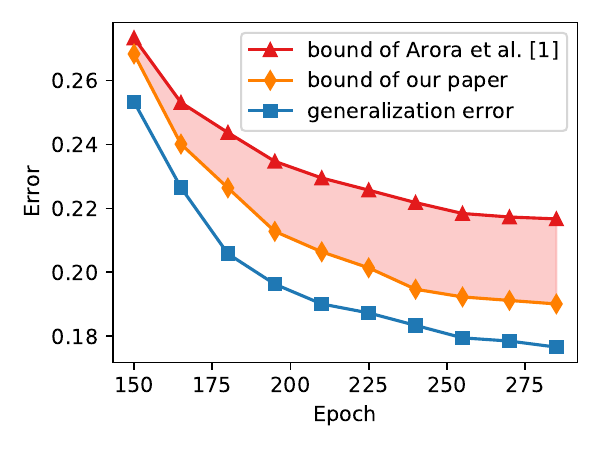}
    \caption{Comparing our bound and [1] to empirical generalization error during training. All bounds are rescaled to be within the same range as the generalization error together.}
    \label{fig:bound}
\end{figure}

\paragraph{Discussion.} The main difference between \citet{arora18compression} and our paper: \citet{arora18compression} proved that the generalization performance of deep neural network is related to the \textit{sparsity} of its parameters, focusing on how to compress the parameters of the trained model. Our paper studies the relationship between DNN generalization performance and \textit{margin distribution} under the condition that DNN parameters is sparsity, focusing on optimizing margin distribution during training. Figure~\ref{fig:bound} shows that the improvement of our bound relative to \citet{arora18compression} (the shaded part in the figure) is because the margin ratio will gradually decrease during the training process. The main difference between \citet{jiang19predicting} and our paper: \citet{jiang19predicting} conjectured that the generalization performance of DNN may be related to the interval distribution. The correlation between generalization and $R^2$ \citep{glantz2001primer} is calculated experimentally, and no theoretical proof is given. Our paper proves theoretically that the generalization performance of DNNs can be bound by the margin ratio and gives the improved algorithm.

\section{Proofs} \label{sec:pr}

In this section, we provide the detailed proofs for the main theorem and lemmas. First, we present a useful lemma as follows:

\begin{lemma}
	\label{extre}
	Let $\mathcal{Q}$ be a probability distribution over the reals. For any random variable $v,v_1,v_2,\dots,v_{m'} \sim \mathcal{Q}$ identically and independently (i.i.d.), we have
	\begin{equation}
		\Pr_{v\sim \mathcal{Q}}\left[v\geq \max_{v_1,\dots,v_{m'}\sim \mathcal{Q}}\{v_1,v_2,\dots,v_{m'}\}\right]= \frac{1}{m'+1}.
	\end{equation}
\end{lemma}

\noindent\emph{Proof of Lemma~\ref{extre}}: Let the Cumulative Distribution Function (CDF) and Probability Density Function (PDF) of random variable $v$ be $F(x)$ and $f(x)$, and we denote the maximum of a set of $m'$ random variables by $v^{(m')}=\max_{v_1,\dots,v_{m'}\sim \mathcal{Q}}\{v_1,v_2,\dots,v_{m'}\}$. Then

\begin{align}
	&\Pr_{v_1,\dots,v_{m'}\sim \mathcal{Q}}\left[v^{(m')}\leq x\right]\\
	=&\Pr_{v_1,\dots,v_{m'}\sim \mathcal{Q}}\left[(v_1\leq x)\wedge \dots \wedge (v_{m'}\leq x )\right]\\
	=&\Pr_{\mathcal{Q}}[v_1\leq x]\times\dots\times\Pr_{\mathcal{Q}}[v_{m'}\leq x]
	=F^{m'}(x).
\end{align}

In other word, the CDF and PDF of the minimum $v^{(m')}$ are $F^{m'}(x)$ and $m'F^{m'-1}(x)f(x)$. Then we can use the minimum value of the sample's set to bound the random variable $v$ with a probability $\frac{m'}{m'+1}$, which converges to $1$ with a rate $\mathcal{O}(1/m')$:

\begin{align}
	&\Pr_{v,v_1,\dots,v_{m'}\sim \mathcal{Q}}\left[v\leq v^{(m')}\right]\\
	=&\Pr_{v,v_1,\dots,v_{m'}\sim \mathcal{Q}}\left[v- v^{(m')}\leq 0\right]\\
	=&\iint_{x\leq y}f(x)F^{m'-1}(y)f(y) \dif x \dif y\\
	=&\int_{-\infty}^{+\infty}m'F^{m'-1}(y)f(y) \dif y \int_{-\infty}^{y} f(x) \dif x\\
	=&\int_{-\infty}^{+\infty}m'F^{m'}(y)f(y) \dif y\label{tg}
	\\
	=&m'F^{m'+1}(y)\left.\right|_{-\infty}^{+\infty}-\int_{-\infty}^{+\infty}m'^2F^{m'}(y)f(y) \dif y \label{th}.
\end{align}

According to Eq.~\eqref{tg} and Eq.~\eqref{th}, we have

\begin{equation}
	\Pr_{v,v_1,\dots,v_{m'}\sim \mathcal{Q}}\left[v\leq v^{(m')}\right]=\frac{m'}{m'+1}.
\end{equation}
\qed

\subsection{Proof of Lemma~\ref{Lemma 1}}

We begin with a lemma as follows:

\begin{lemma}\label{cor:1}
	For any layer $i$, the point-wise \emph{compressibility} of the layer-wise parameters can holds with a probability $1-\frac{1}{m'+1}$ over $\vx\in \mathcal{D}$ as follows:
	\begin{align}
		\mu_i\|\mW_i\|_F\|\phi(\vx^{i-1})\|_2 &\leq \|\vx^{i}\|_2,\\
		\mu_{i,j}\|J_{\vx^{i}}^{i,j}\|_F\|\phi(\vx^{i-1})\|_2  &\leq \|\vx^{j}\|_2,\\
		\|M^{i, j}(\vx^{i}+\eta)-J_{\vx^{i}}^{i, j}(\vx^{i}+\eta)\| &\leq \frac{\|\eta\|\|\vx^{j}\|}{\rho_{\delta}\|\vx^{i}\|}\\
		c\|\phi(\vx^{i})\|_2 &\geq \|\vx^{i}\|_2,\\
		\alpha\|\gamma_{h}(\vx,y)\|_2 &\geq \|\vx^d\|_2,
	\end{align}
	where $m'$ is the size of the validation set.
\end{lemma}

\noindent\textbf{Proof of Lemma~\ref{cor:1}}: According the independence between $S$ and $S'$, we can regard the noise-sensitivity parameters $\frac{1}{\mu_i},\frac{1}{\mu_{i,j}}$, $c$ and $\alpha$ as random variables over reals relying on the randomness of variable $\vx \in S'$. Then, the cushion parameters defined in Assumptions~\ref{ass:1}, \ref{ass:2}, \ref{ass:6}, \ref{ass:3} and \ref{ass:4} can be interpreted as choosing the maximum of multiple independent samples. We first prove Lemma~\ref{extre} on the tail of a random variable $v\sim\mathcal{Q}$ by choosing the maximum of multiple independent samples of the random variable. Specifically, using the following simple lemma based on the distribution of the maximum, we can guarantee the point-wise \emph{compressibility} of the learned parameters over the underlying data distribution $\mathcal{D}$ with a high probability by calculating the maximum of the empirical dataset, i.e., $\frac{1}{\mu_i}, \frac{1}{\mu_{i,j}}$, $c$ and $\alpha$. 
\qed\\

\noindent\textbf{Proof of Lemma~\ref{Lemma 1}}: First, we need to bound the perturbation of linear operator caused by injecting a scaled Gaussian noise $\mU=\beta\|\mW\|_F, \mathbb{E}[\beta\beta^\top]=\sigma\mI$. For any fixed vectors $\va,\vb$, we have
\begin{align}
	&\mathbb{E}_{\beta} \|\va^\top (\mW+\mU) \vb - \va^\top \mW \vb \|_2= \mathbb{E}_{\beta}\|\vb \|_2\|\va^\top \mU\mU^\top\va \|_2\notag\\ &= \mathbb{E}_{\beta}\|\mW\|_F\|\vb \|_2\|\va^\top \beta\beta^\top\va \|_2 \\&= \sigma\|\mW\|_F\|\va\|_2\|\vb\|_2.
\end{align}
According the Markov inequality, we have
\begin{align}\label{eq:27}
	&\Pr_{\beta}\left[ \|\va^\top (\mW+\mU) \vb - \va^\top \mW \vb \|_2 \geq \sigma\sqrt{d}/\sqrt{\delta} \|\mW\|_F\|\va\|_2\|\vb\|_2\right]\notag\\&\leq \frac{\sigma^2\|\mW\|_F^2 \|\va\|^2_2\|\vb\|^2_2}{(d\sigma^2/\delta)\|\mW\|_F^2\|\va\|^2_2\|\vb\|^2_2} =\frac{\delta}{d}.
\end{align}

Now, we will bound the perturbation of the $d$-layer deep nets by induction. For any layer $i\geq0$, let $x^j$ be the output at layer $j$ with original net and $\hat{\vx}_i^j$ be the output at layer $j$ if the weights $\mW_1,\dots,\mW_i$ in the first layers are replaced with $\mW_1+\mU_1,\dots,\mW_i+\mU_i$. The induction hypothesis is then following:

Consider any $0<\epsilon\leq1$, the following is true with probability $1-\frac{i\delta}{d}$ over $\mW_1+\mU_1,\dots,\mW_i+\mU_i$ for any $j\geq i$:
\begin{equation}
	\|\hat{\vx}_{i}^{j}-\vx^{j}\|_2^2 \leq\sum_{l=1}^{i}\frac{c^2 d \sigma^2}{\delta\mu_{l}^2\mu_{l \rightarrow}^2}\|\vx^{j}\|_2^2.
\end{equation}

For the base case $i = 0$, since we are not perturbing the input, the inequality is trivial. Now assuming that the induction hypothesis is true for $i-1$, we consider what happens at layer $i$.
\begin{align}
	\label{eq:29} \|\hat{\vx}_{i}^{j}-\vx^{j}\|_2^2&=\|(\hat{\vx}_{i}^{j}-\hat{\vx}_{i-1}^{j})+(\hat{\vx}_{i-1}^{j}-\vx^{j})\|_2^2\notag \\&\leq2\|(\hat{\vx}_{i}^{j}-\hat{\vx}_{i-1}^{j})\|_2^2+2\|\hat{\vx}_{i-1}^{j}-\vx^{j}\|_2^2
\end{align}

The second term in Eq.~\eqref{eq:29} can be bounded by $\sum_{l=1}^{i-1}\frac{c^2  \sigma^2}{\mu_{l}^2\mu_{l \rightarrow}^2}\|\vx^{j}\|_2^2$ by induction hypothesis. Therefore, it is enough to show that the first term in Eq.~\eqref{eq:29} is bounded by $\frac{c^2  \sigma^2}{\mu_{i}^2\mu_{i \rightarrow}^2}\|\vx^{j}\|_2^2$. We decompose the error into two error terms one of which corresponds to the error propagation through the network if activation were fixed and the other one is the error caused by change in the activations:
\begin{align}
	& \| (\hat{\vx}_{i}^{j}-\hat{\vx}_{i-1}^{j} ) \|\\&= \|M^{i, j} ((\mW_i+\mU_i) \phi (\hat{\vx}^{i-1} ) )-M^{i, j} (\mW_i \phi (\hat{\vx}^{i-1} ) ) \|\\
	&= \|M^{i, j} ((\mW_i+\mU_i) \phi (\hat{\vx}^{i-1} ) )-M^{i, j} (\mW_i \phi (\hat{\vx}^{i-1} ) )\notag\\&\quad+J_{\vx^{i}}^{i, j} (\mU^{i} \phi (\hat{\vx}^{i-1} ) )-J_{\vx^{i}}^{i, j} (\mU^{i} \phi (\hat{\vx}^{i-1} ) ) \|\\
	\label{eq:32}&\leq \|J_{\vx^{i}}^{i, j} (\mU^{i} \phi (\hat{\vx}^{i-1} ) ) \|+ \|M^{i, j} ((\mW_i+\mU_i) \phi (\hat{\vx}^{i-1} ) )\notag\\&-M^{i, j} (\mW_i \phi (\hat{\vx}^{i-1} ) )-J_{\vx^{i}}^{i, j} (\mU^{i} \phi (\hat{\vx}^{i-1} ) ) \|.
\end{align}

The first term in Eq.~\eqref{eq:32} is bounded by:

\begin{align}
	& \|J_{\vx^{i}}^{i, j} \mU^{i} \phi (\hat{\vx}^{i-1} ) \|_2\label{44}\\
	&\leq (\sqrt{d}\sigma/\sqrt{6\delta}) \|J_{\vx^{i}}^{i, j} \|_2 \|\mW_i \|_{F} \|\phi (\hat{\vx}^{i-1} ) \|_2 \label{lip}\\
	&\leq (\sqrt{d}\sigma/\sqrt{6\delta}) \|J_{\vx^{i}}^{i, j} \|_2 \|\mW_i \|_{F} \|\hat{\vx}^{i-1} \|_2\label{ind}\\
	&\leq (\sqrt{d}\sigma/\sqrt{3\delta}) \|J_{\vx^{i}}^{i, j} \|_2 \|\mW_i \|_{F} \|\vx^{i-1} \|_2\label{act}\\
	&\leq (c\sqrt{d}\sigma/\sqrt{3\delta}) \|J_{\vx^{i}}^{i, j} \|_2 \|\mW_i \|_{F} \|\phi (\vx^{i-1} ) \|_2\label{lay}\\
	&\leq (c\sqrt{d}\sigma/\sqrt{3\delta}\mu_{i}) \|J_{\vx^{i}}^{i, j} \|_2 \|\mW_i \phi (\vx^{i-1} ) \|_2\\
	&\leq (c\sqrt{d}\sigma/\sqrt{3\delta}\mu_{i}\mu_{i \rightarrow}) \|\vx^{j} \|_2\label{int}.
\end{align}

where Equation~\eqref{44} is bounded by Eq.~\eqref{eq:27}, Equation~\eqref{lip} is bounded by Lipschitzness of the activation function, Equation~\eqref{ind} is bounded by inductive hypothesis, Equation~\eqref{act} is bounded by activation contraction, Equation~\eqref{lay} is bounded by layer cushion, and Equation~\eqref{int} is bounded by interlayer cushion.

The second term in Eq.~\eqref{eq:32} can be bounded as:
\begin{align}
	& \|M^{i, j} ((\mW_i+\mU_i) \phi (\hat{\vx}^{i-1} ) )-M^{i, j} (\mW_i \phi (\hat{\vx}^{i-1} ) )\notag\\&\quad-J_{\vx^{i}}^{i, j} (\mU^{i} \phi (\hat{\vx}^{i-1} ) ) \|_2\\
	&= \| (M^{i, j}-J_{\vx^{i}}^{i, j} ) ((\mW_i+\mU_i) \phi (\hat{\vx}^{i-1} ) )\notag\\&\quad- (M^{i, j}-J_{\vx^{i}}^{i, j} ) (\mW_i \phi (\hat{\vx}^{i-1} ) ) \|_2\\
	\label{eq:42} &= \| (M^{i, j}-J_{\vx^{i}}^{i, j} ) ((\mW_i+\mU_i) \phi (\hat{\vx}^{i-1} ) ) \|_2\notag\\&\quad+ \| (M^{i, j}-J_{\vx^{i}}^{i, j} )(\mW_i \phi(\hat{\vx}^{i-1})) \|_2.
\end{align}

Both terms in Eq.~\eqref{eq:42} can be bounded using Assumption~\ref{ass:6}. By notations we find $\mW^{i} \phi(\hat{\vx}^{i-1})=\hat{x}_{i-1}^{i}$. By induction hypothesis, we have that $\|\mW^{i} \phi(\hat{\vx}^{i-1})-\vx^{i}\|_2^2 \leq\sum_{l=1}^{i-1}\frac{c^2 d \sigma^2}{\delta\mu_{l}^2\mu_{l \rightarrow}^2}\|\vx^{i}\|_2^2$. Now by interlayer smoothness property, $\|(M^{i, j}-J_{\vx^{i}}^{i, j})(\mW^{i} \phi(\hat{\vx}^{i-1})\|_2^2 \leq \frac{\sum_{l=1}^{i-1}\frac{c^2 d \sigma^2}{\delta\mu_{l}^2\mu_{l \rightarrow}^2}\|\vx^{j}\|}{\rho_{\delta}} \leq(\sum_{l=1}^{i-1}\frac{c^2d \sigma^2}{\delta\mu_{l}^2\mu_{l \rightarrow}^2} )\| \vx^{j} \|_2^2/(3d)\simeq\frac{i-1}{3d}\frac{c^2 d \sigma^2}{\delta\mu_{i}^2\mu_{i \rightarrow}^2}\|\vx^{j}\|_2^2$. Similar to this term, $\|(M^{i, j}-J_{\vx^{i}}^{i, j})((\mW^{i}+\mU_i) \phi(\hat{\vx}^{i-1}))\| \leq(\sum_{l=1}^{i}\frac{c^2 d\sigma^2}{\delta\mu_{l}^2\mu_{l \rightarrow}^2} )\| \vx^{j} \|/(3d)\simeq\frac{i}{3d}\frac{c^2 d \sigma^2}{\delta\mu_{i}^2\mu_{i \rightarrow}^2}\|\vx^{j}\|_2^2$. Putting everything together completes the induction with probability at least $1-\delta$ (if $i=d$).

Instead of assuming that the input domain $\mathcal{X}$ is bounded by a constant $B$, we assume that the input boundary is relative to the expected value which implies the data-distribution information: $\max_{\vx}\|\vx^d\|_2 \leq \mathcal{O}(\mathbb{E}_{\mathcal{D}}\|\vx^d\|_2)$. According to the margin contraction property, we can use the first- and second-statistics of the margin in the last layer $\mathbb{E}_{\mathcal{D}} [\gamma_{h}(\vx,y)]=r, \Var_{\mathcal{D}}[\gamma_{h}(\vx,y)]=\theta^2$ to bound the perturbation instead of the worst situation:
\begin{align}
	\max_{\vx}\|\vx^d\|_2^2 &\leq \mathcal{O}(\mathbb{E}_{\mathcal{D}}\|\vx^d\|_2^2) \leq \mathcal{O}(\alpha^2 \mathbb{E}_{\mathcal{D}}\|\gamma_{h}(\vx,y)\|_2^2) \\&\leq \mathcal{O}(\alpha^2(r^2+\theta^2)) \leq \mathcal{O}(\alpha^2\left(r + \theta\right) ^2)
\end{align}

Connecting these two inequalities we prove that the equality holds with a probability at least $1/2$:
\begin{equation}
	|f_{\vw+\vu}(\vx)-f_{\vw}(\vx)|_2^2 \leq \mathcal{O}\left(
	\sum_{i=1}^{d}\frac{d\alpha^2 c^2(r+\theta)^2\sigma^2}{\mu_{i}^2\mu_{i \rightarrow}^2}\right).
\end{equation}
\qed\\

\subsection{Proof of Lemma~\ref{Lemma 2}}\label{sec:4.2}

\noindent\textbf{Proof of Lemma~\ref{Lemma 2}}:
Let $\vw'=\vw+\vu$, Let $\mathcal{S}_{\vw}$ be the set of perturbations with the following property:
\begin{equation}
	\mathcal{S}_{\vw} \subseteq\left\{\vw^{\prime}\bigg|\max _{\vx \in \mathcal{X}}| f_{\vw^{\prime}}(\vx)-f_{\vw}(\vx)|_{2}<\frac{r-\theta}{8\sqrt{\rho}}\right\},
\end{equation}
then we will have $\max _{\vx \in \mathcal{X}}| f_{\vw^{\prime}}(\vx)-f_{\vw}(\vx)|_{\infty}<\sqrt{\rho}\max _{\vx \in \mathcal{X}}| f_{\vw^{\prime}}(\vx)-f_{\vw}(\vx)|_{2}<\frac{r-\theta}{8}$.

Let $q$ be the probability density function over the parameters $\vw'$. We construct a new distribution $\tilde{Q}$ over predictors $f_{\tilde{\vw}}$ where $\tilde{\vw}$ is restricted to $\mathcal{S}_{\vw}$ with the probability density function:
\begin{equation}
	\tilde{q}(\tilde{\vw})=\frac{1}{Z} 
	\begin{cases}
		q(\tilde{\vw}) & \tilde{\vw} \in \mathcal{S}_{\vw} \\ 0 & \text { otherwise }
	\end{cases}
\end{equation}
According to the lemma assumption, we have $Z=\mathbb{P}\left[\vw^{\prime} \in \mathcal{S}_{\vw}\right] \geq \frac{1}{2}$. Therefore, we can bound the change of the margins for any instance:
\begin{equation}
	\max _{i, j \in[k], \vx \in \mathcal{X}}\left|\left(\left|f_{\tilde{\vw}}(\vx)[i]-f_{\tilde{\vw}}(\vx)[j]\right|\right)-\left(\left|f_{\vw}(\vx)[i]-f_{\vw}(\vx)[j]\right|\right)\right|<\frac{r-\theta}{2}
\end{equation}
Here we define a perturbed loss function as:
\begin{equation}
	L'_{r,\theta}(h) = \Pr_{\mathcal{D}} \left[\gamma_{h}(\vx,y) \leq \frac{r - \theta}{2}   \right] + \Pr_{\mathcal{D}} \left[ \gamma_{h}(\vx,y)   > r + \theta + \frac{r - \theta}{2}  \right] .
\end{equation}
We can get the following:
\begin{align}
		&L_{0}\left(f_{\vw}\right) \leq L'_{r,\theta}\left(f_{\tilde{\vw}}\right) \\
		&\widehat{L}'_{r,\theta}\left(f_{\tilde{\vw}}\right) \leq \widehat{L}_{r,\theta}\left(f_{\vw}\right)
\end{align}
Finally, using the proof of \citet[Lemma 1]{neyshabur18spectrally}, with probability $1-\delta$ over the training set we have:
\begin{align}
	L_{0}\left(f_{\vw}\right) & \leq \mathbb{E}_{\tilde{\vw}}\left[L'_{r,\theta}\left(f_{\tilde{\vw}}\right)\right] \\
	& \leq \mathbb{E}_{\tilde{\vw}}\left[\widehat{L}'_{r,\theta}\left(f_{\tilde{\vw}}\right)\right]+2 \sqrt{\frac{2\left(\KL(\tilde{\vw} \| P)+\ln \frac{2 m}{\delta}\right)}{m-1}} \\
	& \leq \widehat{L}_{r,\theta}\left(f_{\vw}\right)+2 \sqrt{\frac{2\left(\KL(\tilde{\vw} \| P)+\ln \frac{2 m}{\delta}\right)}{m-1}} \\
	& \leq \widehat{L}_{r,\theta}\left(f_{\vw}\right)+4 \sqrt{\frac{\KL\left(\vw^{\prime} \| P\right)+\ln \frac{6 m}{\delta}}{m-1}}
\end{align}

\qed\\

\subsection{Proof of Theorem~\ref{theorem 1}}\label{sec:4.3}
\noindent\textbf{Proof of Theorem~\ref{theorem 1}}:
Since Lemma~\ref{Lemma 1} proves that the perturbation caused by random vector $\vu$ is bounded by a term relative to the variance $\sigma$, we can preset the value of $\sigma$ to make the random perturbation satisfy the condition for Lemma~\ref{Lemma 2}. Bounding the Kullback-Leibler divergence term by $\|\vw\|_2^2/\|\vu\|_2^2$ in PAC-Bayesian theorem, we can attain the generalization bound based on a specific margin distribution.

The proof involves chiefly two steps. In the first step we bound the maximum value of perturbation of parameters to satisfy the condition that the change of output restricted by hyper-parameters of margin $r$ and $\theta$, using Lemma~\ref{Lemma 1}. In the second step we prove the final margin generalization bound through Lemma~\ref{Lemma 2} with the value of Kullback-Leibler divergence term calculated based on the bound in the first step.
\begin{align*}
	|f_{\vw+\vu}(\vx)-f_{\vw}(\vx)|_2^2 &\leq \mathcal{O}\left(
	\sum_{i=1}^{d}\frac{d\alpha^2c^2(r+\theta)^2\sigma^2}{\mu_{i}^2\mu_{i \rightarrow}^2}\right)\\
	&= (\frac{r-\theta}{8\sqrt{\rho}})^2
\end{align*}
We can derive $\sigma=\frac{r-\theta}{8\alpha cd\sqrt{\rho}(r+\theta)\sqrt{\sum_{i=1}^{d}\frac{1}{\mu_{i}^2\mu_{i \rightarrow}^2}}}$ from the above inequality. Naturally, we can calculate the Kullback-Leibler divergence in Lemma \ref{Lemma 1} with the chosen distributions for $P\sim \mathcal{N}(0,\sigma^2\mathbf{I})$.
\begin{align}
	\KL(\vw+\vu\|P)&\leq\frac{|\vw|^2}{2|\vw|^2|\eta\eta^\top|^2}=\frac{1}{2\sigma^2}\\
	&\leq\mathcal{O}\left( \frac{(r+\theta)^2}{(r-\theta)^2}\sum_{i=1}^{d}\frac{d\rho\alpha^2c^2}{\mu_{i}^2\mu_{i \rightarrow}^2}  \right)
\end{align}	
Put it in Lemma~\ref{Lemma 2} and let $\lambda=\theta/r$, with probability at least $1-\delta$ and for all $\vw$ such that, we have:
\begin{equation}
	L_0(h) \leq \widehat{L}_{r,\theta}(h) + \mathcal{O} \left( \sqrt{\frac{\frac{(1+\lambda)^2}{(1-\lambda)^2} \sum_{i=1}^{d}\frac{d\rho\alpha^2c^2}{\mu_{i}^2\mu_{i \rightarrow}^2}+\ln\frac{dm}{\delta}}{m}} \right).
\end{equation}
\qed

\section{Optimizing margin distribution measure}\label{method}

The generalization theory shows the importance of optimizing the margin distribution ratio $\lambda$. The result inspires us to find a margin distribution band ($r-\theta\leq \gamma_{h}(\vx,y)<r+\theta$) containing as many training samples as possible to minimize the empirical estimate loss $\widehat{L}_{r,\theta}$, but also a ratio $\lambda=\theta/r$ as small as possible to minimize the generalization gap $L_0(h)-\widehat{L}_{r,\theta}(h)$. This type of loss function was first proposed by \citet{zhang16odm} to optimize the first- and second-order statistics of margin distribution. We formulate a convex margin distribution loss function for DNNs:

\begin{definition}(Convex margin distribution loss function).
	For a labeled sample $(\vx,y)\in\mathcal{D}$, we denote its margin by $\gamma_h$ which is defined as Eq.~\eqref{eq:margin}. We define the margin distribution loss for networks (mdNet loss) as: 
	\begin{equation} \label{loss_fn}
		\ell_{r,\theta,\eta}(h(\vx),y) =
		\begin{cases}
			\frac{(r-\theta-\gamma_h)^2}{(r-\theta)^2} & \gamma_h \leq r - \theta\\
			0 & r - \theta < \gamma_h \leq r + \theta\\
			\frac{\eta (r + \theta - \gamma_h)^2}{(r + \theta)^2} & \gamma_h > r + \theta,
		\end{cases}
	\end{equation}
	where $r$ is the margin mean parameter, $\theta$ is the margin variance parameter and $\eta$ is a parameter to trade off two different kinds of deviation (keeping the balance on both sides of the margin mean). Figure~\ref{fig:convex_loss} shows the shape of this convex loss function.
\end{definition}

Equation~\eqref{loss_fn} will produce a square loss when the margin satisfies $\gamma_h \leq r-\theta$ or $\gamma_h \geq r + \theta$. Therefore, our margin loss function will force the zero-loss band to contain as many sample points as possible. The ratio of hyper-parameters $\lambda=\theta/r$ can control the capacity measure, which implies our measure is dependent to our specific learning algorithm (loss function with specific hyper-parameters). Since our loss function aims at finding a decision boundary which is determined by the entire margin distribution, instead of the minority samples that have minimum margins, we name our method as \underline{m}argin \underline{d}istribution \underline{Net}works (mdNet).

\section{Experiments} \label{experiment}

In Subsection~\ref{7.1}, we introduce the configuration of datasets and models. In Subsection~\ref{7.2}, We design an ablation experiment to verify the superiority of our method. In Subsection~\ref{7.3}, we show the correlation between separability of representations and margin ratio via visualization. In Subsection~\ref{7.4}, we design experiments to confirm that our method can control the capacity of deep nets. In Subsection~\ref{sec:hyper}, we discuss the influence of the different hyper-parameters on the test accuracy. 

\subsection{Configuration}\label{7.1}

Since our method only works on the loss function part of deep models and does not change the architecture of deep neural networks, we can verify the effectiveness of mdNet on the classic CNNs (convolutional neural networks) and image classification benchmark datasets. We consider the following architectures and datasets: a LeNet architecture for MNIST dataset \citep{lecun98gradient}, an AlexNet architecture \citep{alex12imagenet} for CIFAR-10 dataset \citep{krizhevsky09learning} and a ResNet-18 architecture \citep{he16res} for ImageNet dataset \citep{deng2012imagenet}. From the literature, these datasets come pre-divided into training and testing sets, therefore in our experiments, we use them in their original format. The loss functions used for comparison in the experiments are as follows: cross-entropy loss (abbr., xent), hinge loss and soft hinge loss. Hinge loss \citep{cortes95svm} and soft hinge loss \citep{liu16large} are loss functions specially proposed to optimize the minimum margin, both of them are inspired the traditional margin theory.

As for details about the architecture, we remove the weight decay \citep{krogh92simple}, dropout \citep{srivastava14dropout} and batch normalization (BN) \citep{ioffe15batch} from all the models, because the batch normalization operation and weight decay will shift the data distribution. The notable dropout technique, in which some neurons are dropped from the DNNs in each iteration, can also be viewed as an ensemble method composed of different neural networks, with different dropped neurons \citep{baldi13understanding}. It is hard to analyze the influence of the ensemble structure on the margin distribution, so we remove this technique in these architectures in the experiments except to understand the contribution of the components to the whole models in the ablation study.

For \textit{special hyper-parameters}, including the expected margin parameter and margin variance parameter for mdNet loss model, and margin parameter for hinge loss model, we perform hyper-parameter search. We hold out 5000 samples of the training set as a validation set, and use the remaining samples to train models with different special hyper-parameters values on all datasets. As for the common hyper-parameters, such as learning rate and momentum, we set them as the default commonly used values in PyTorch \citep{paszke2017automatic} for all the models. We chose batch stochastic gradient descent as the optimizer. We run all the experiments on four K80 GPU machines. As for the influence of the different hyper-parameters on the test accuracy, we discuss it empirically in Subsection~\ref{sec:hyper}.


\subsection{Ablation study} \label{7.2}

Since the optimization algorithm proposed in this paper only focuses on the improvement of loss function, we design an ablation experiment to study the performance of our proposed mdnet method and the traditional benchmark loss functions under different regularizations in Table~\ref{tab:ablation}. The mdNet loss outperforms the others consistently across different situations, no matter whether dropout, batch normalization or the entire dataset are used or not. The experiments are evaluated on three MNIST (LeNet), CIFAR-10 (AlexNet) and ImageNet (ResNet-18) datasets. Specifically, when the amount of training samples is small, the advantage of mdNet loss is significant. Moreover, the mdNet loss function can cooperate with both batch normalization and dropout, achieving the best performance in Table~\ref{tab:ablation}, which is highlighted in bold red text. Unlike dropout and batch normalization which lack solid theoretical grounds, the mdNet loss function is inspired by the margin distribution bound in Theorem~\ref{theorem 1}, which guides us to find a suitable margin ratio to restrict the model capacity and alleviate the overfitting problem efficiently.

\subsection{Feature visualization} \label{7.3}

\begin{figure}[b]
	\centering
	\subfigure[The t-SNE visualization of learned representations of different models for MNIST.]{\label{fig:tsne_a}
	\begin{minipage}[]{0.95\textwidth}
		\centering
		\includegraphics[width=\textwidth]{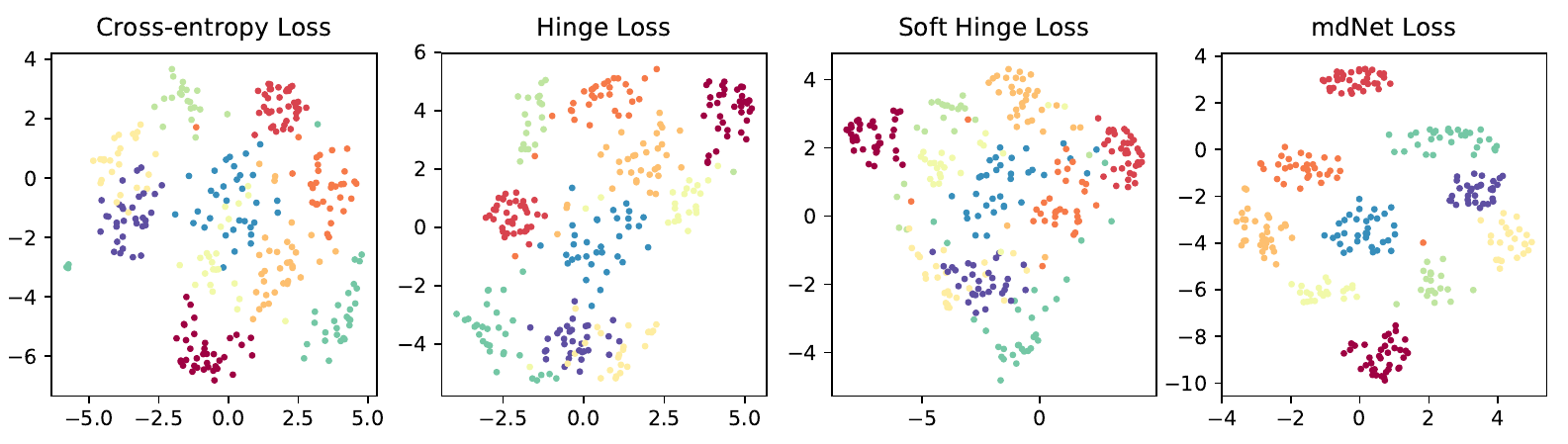}
	\end{minipage}}
	\hfill
	\subfigure[The t-SNE visualization of learned representations of different models for CIFAR-10.]{\label{fig:tsne_b}
	\begin{minipage}[]{0.95\textwidth}
		\centering
		\includegraphics[width=\textwidth]{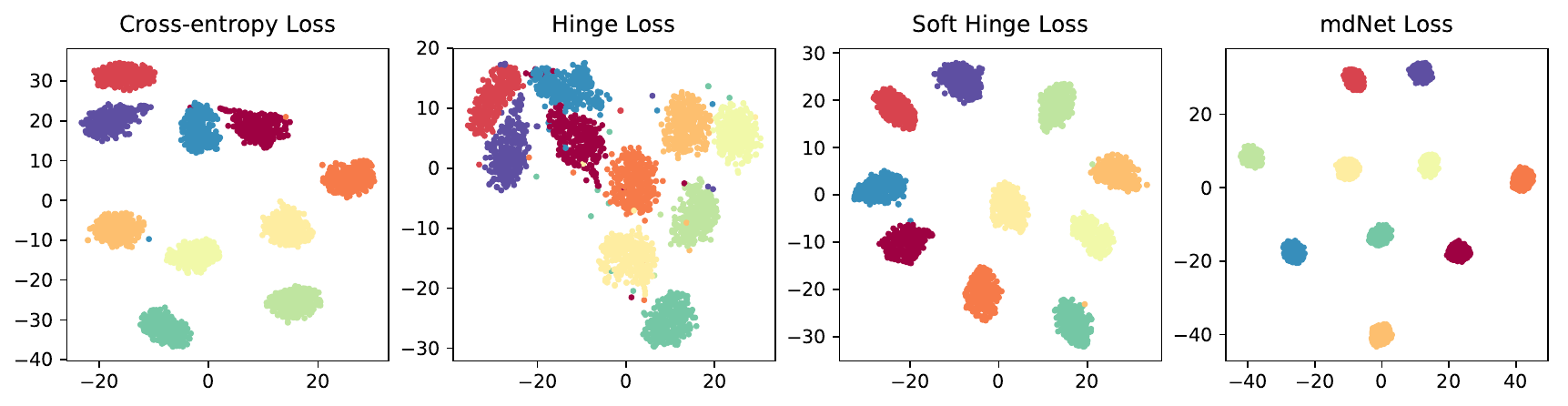}
	\end{minipage}}
	\hfill
	\subfigure[The t-SNE visualization of learned representations of different models for ImageNet.]{\label{fig:tsne_resnet}
	\begin{minipage}[]{0.95\textwidth}
		\centering
		\includegraphics[width=\textwidth]{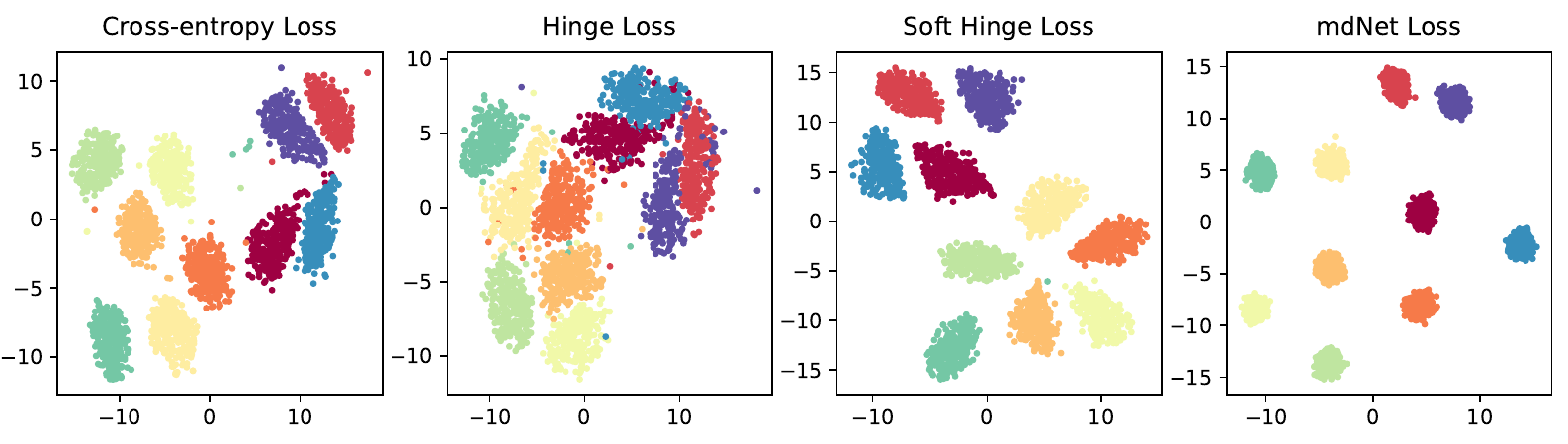}
	\end{minipage}}
	\hfill
	\vskip 0.2in
	\subfigure[The variance decomposition of learned representations of different models for MNIST.]{\label{fig:tsne_c}
	\begin{minipage}[]{0.30\textwidth}
		\centering
		\includegraphics[width=\textwidth]{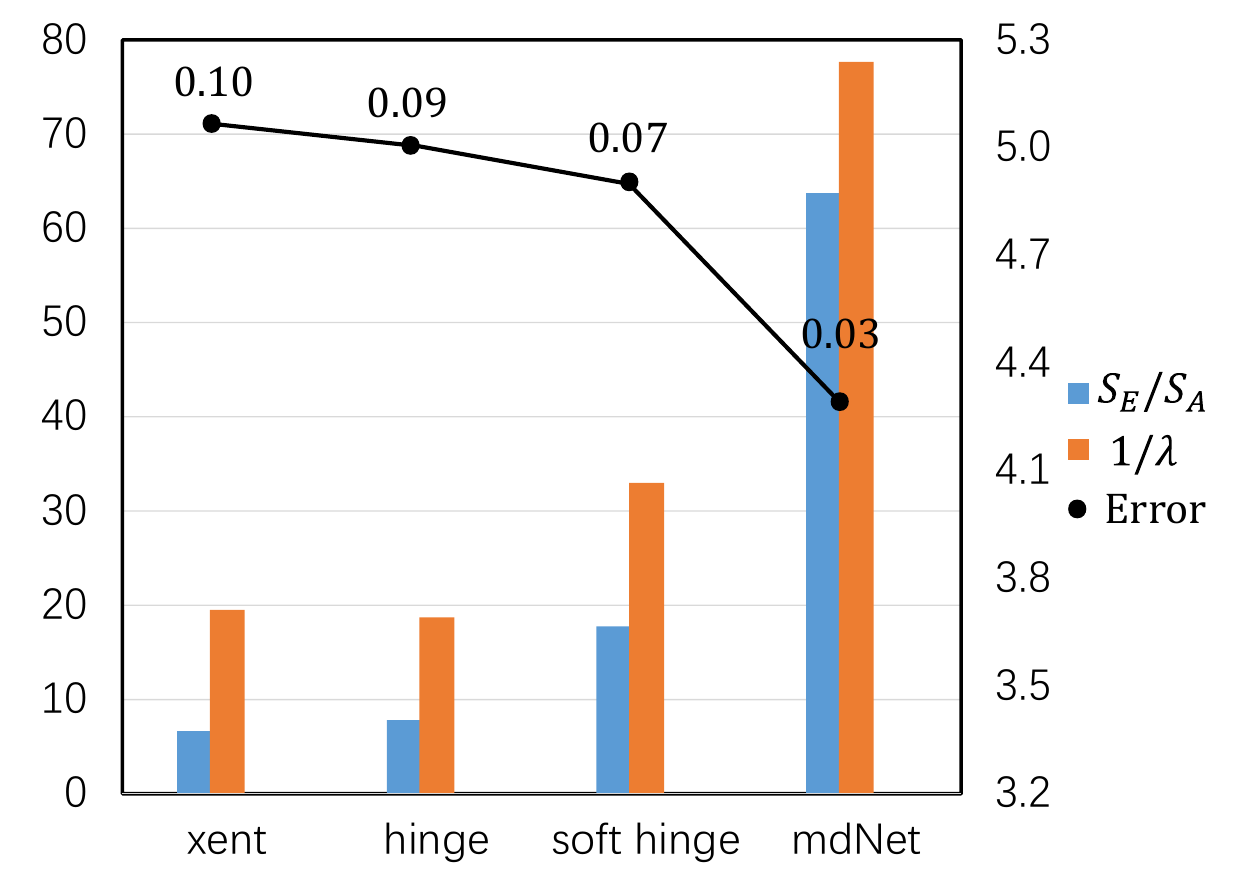}
	\end{minipage}}
	\hfill
	\subfigure[The variance decomposition of learned representations of different models for CIFAR-10.]{\label{fig:tsne_d}
	\begin{minipage}[]{0.30\textwidth}
		\centering
		\includegraphics[width=\textwidth]{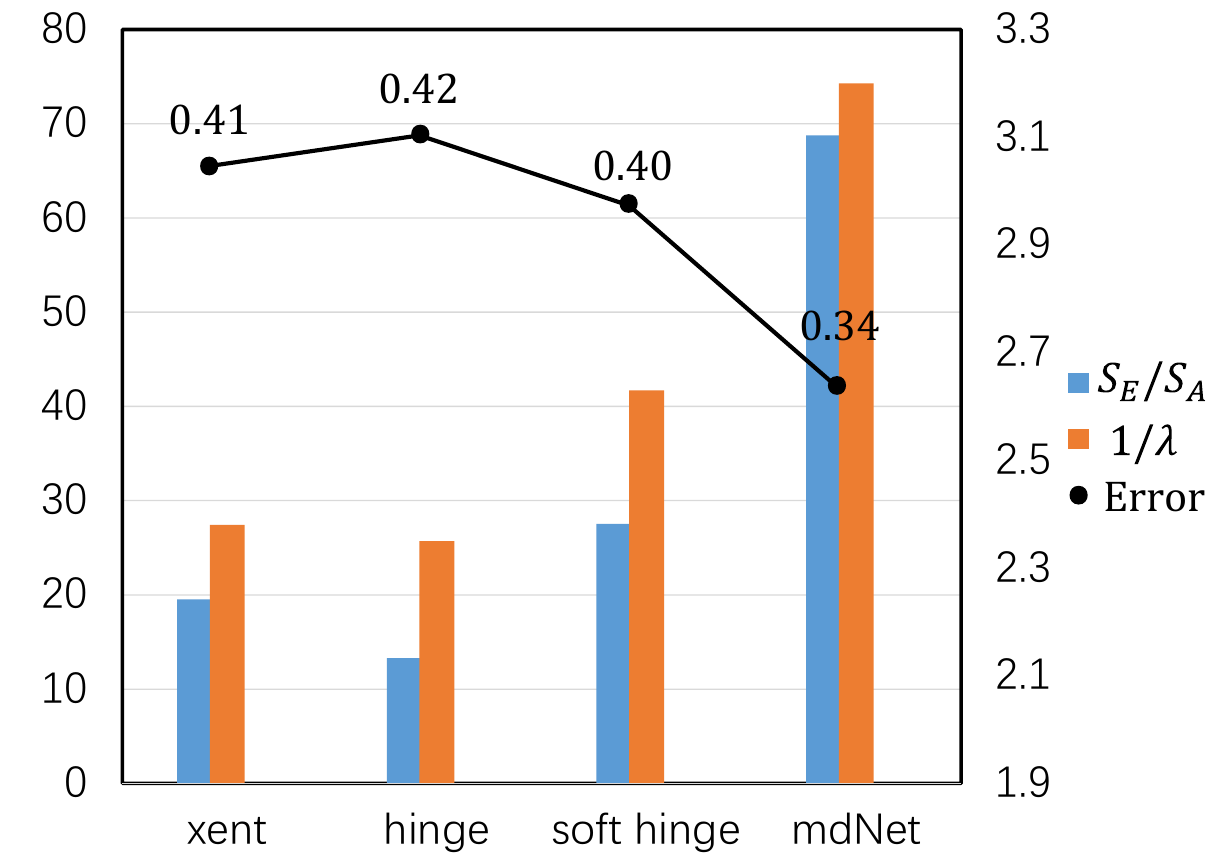}
	\end{minipage}}
	\hfill
	\subfigure[The variance decomposition of learned representations of different models for ImageNet.]{\label{fig:tsne_e}
		\begin{minipage}[]{0.30\textwidth}
			\centering
			\includegraphics[width=\textwidth]{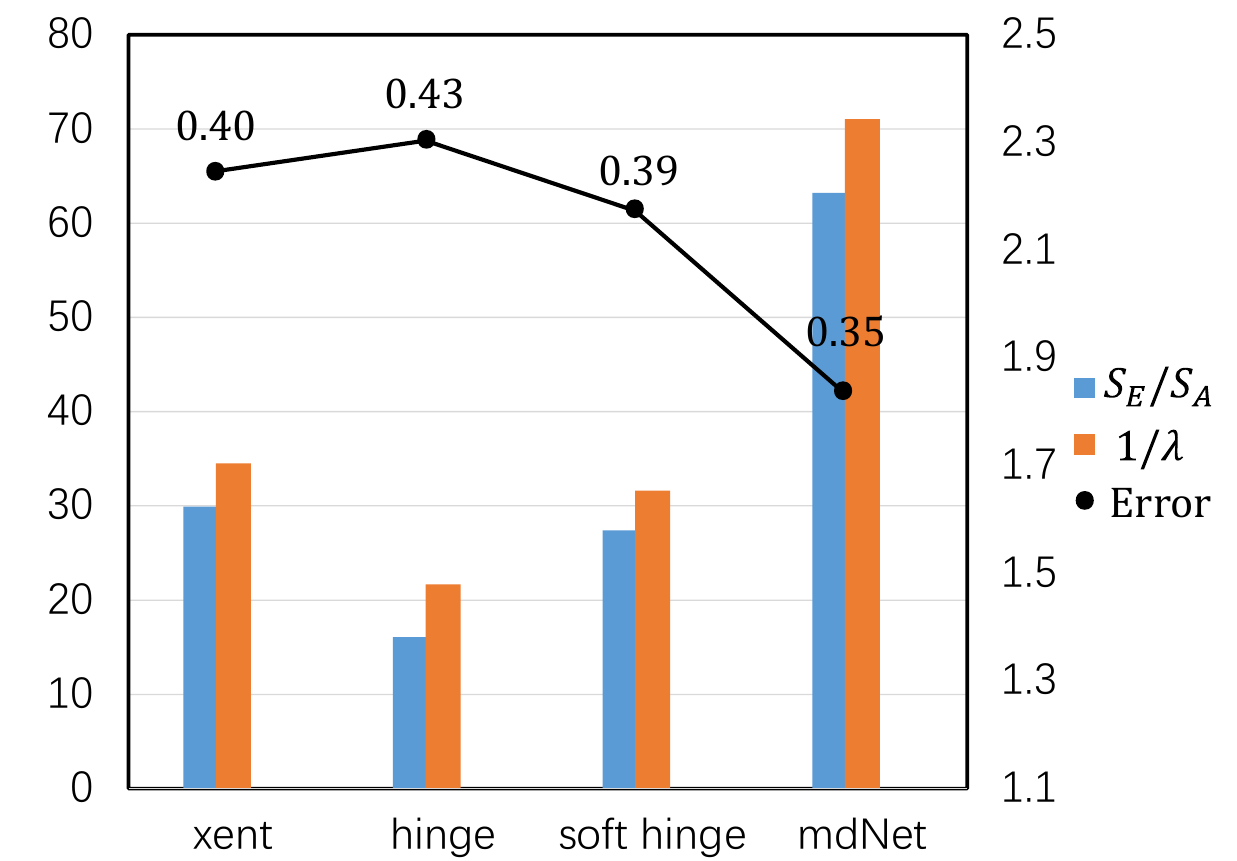}
	\end{minipage}}
	\caption{The quality of feature representations generated by different models on the MNIST, CIFAR-10 and ImageNet datasets.}
	\label{fig:tsne}
	\vskip -0.0in
\end{figure}

\begin{figure}[b]
	\centering
	\subfigure[LeNet for MNIST.]{\label{fig:curve1_a}
		\begin{minipage}[]{0.30\textwidth}
			\centering
			\includegraphics[width=\textwidth]{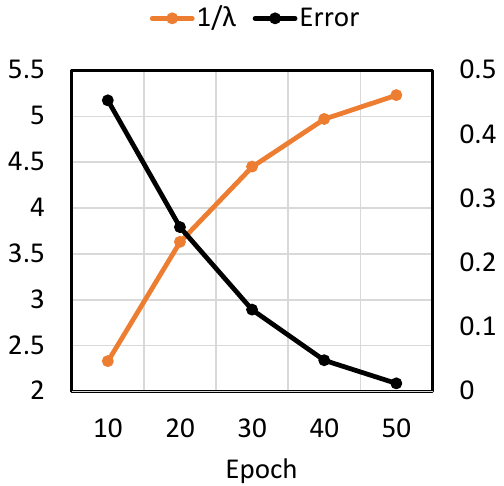}
	\end{minipage}}
	\hfill
	\subfigure[AlexNet for CIFAR-10.]{\label{fig:curve1_b}
		\begin{minipage}[]{0.30\textwidth}
			\centering
			\includegraphics[width=\textwidth]{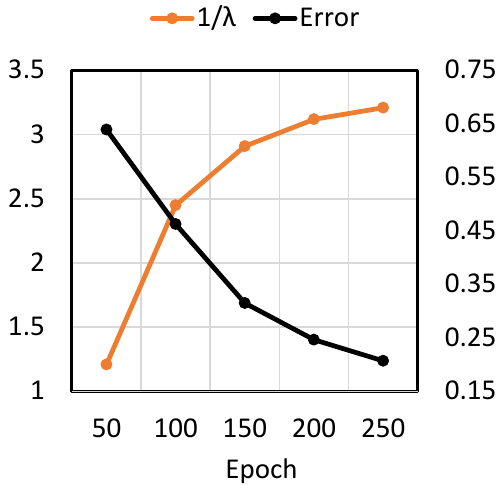}
	\end{minipage}}
	\hfill
	\subfigure[ResNet-18 for ImageNet.]{\label{fig:curve1_c}
		\begin{minipage}[]{0.30\textwidth}
			\centering
			\includegraphics[width=\textwidth]{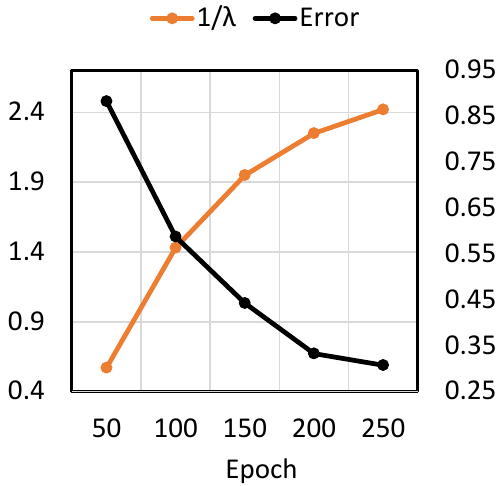}
	\end{minipage}}
	\caption{Test error and margin ratio across epochs on mdNet models for MNIST, CIFAR-10 and ImageNet datasets.}
	\label{fig:curve1}
\end{figure}

In this experiment, we use t-SNE method to visualize the learned representations on the last hidden layer. Figure~\ref{fig:tsne_a},~\ref{fig:tsne_b} and~\ref{fig:tsne_resnet} plot the 2D t-SNE \citep{maaten08visualizing} embedding image on limited datasets, including MNIST (LeNet), CIFAR-10 (AlexNet) and 10-class ImageNet (ResNet-18). Consistently, we can find that the result of mdNet loss model is better than all the others, the distribution of samples which have the same label are more compact. To quantify the degree of separability of data distribution, we perform a variance decomposition on the data in the embedding space. By comparing the ratio of inter-class variance $S_E$ to intra-class variance $S_A$ in Figure~\ref{fig:tsne_c},~\ref{fig:tsne_d} and~\ref{fig:tsne_e}, we see that the mdNet loss always attain the most separable distribution among these four loss functions. Moreover, the visualization result is consistent with the margin distribution ratio $1/\lambda$ of these four models, which means that optimizing the margin distribution (searching an appropriate margin ratio $\lambda$) is helpful to attain a good learned representation space. This representation features space can further alleviate the overfitting problem of deep learning, we verify empirically that a network trained with mdNet loss shows stronger clustering. Specially, Figure~\ref{fig:tsne_c},~\ref{fig:tsne_d} and~\ref{fig:tsne_e} show the relationship between the margin ratio and test error. Moreover, Figure~\ref{fig:curve1} plots the test error of mdNet and the margin ratio across the different epochs. We can see that more compact margin distribution gets better prediction performance across different models and epochs. This exhibits that optimizing margin distribution can indeed improve the learning ability of deep nets.

\subsection{Controlling capacity} \label{7.4}

The first two experiments have demonstrated that our mdNet can outperform other classical loss functions and our method can learn a more separable feature representation, as the corresponding margin ratio is also smaller. However, it leads to the last question:

 \textit{Can smaller margin ratio reduce the capacity of models and accelerate the convergence of generalization gap?}\\

Let's go back to the theoretical result obtained by Theorem~\ref{theorem 1}. The generalization gap of the model is bounded by $\Lambda_{\lambda,\vw}/\sqrt{m}$, where the margin distribution measure $\Lambda_{\lambda,\vw}\propto\frac{1+\lambda}{1-\lambda}$ determines the worse case of generalization gap when the number of samples are equal:
\begin{equation}
	\mathrm{Generalization\ Gap} \leq \mathcal{O}(\frac{\Lambda_{\lambda,\vw}}{\sqrt{m}}).
\end{equation}

\begin{figure}[b]
	\centering
	\includegraphics[width=\textwidth]{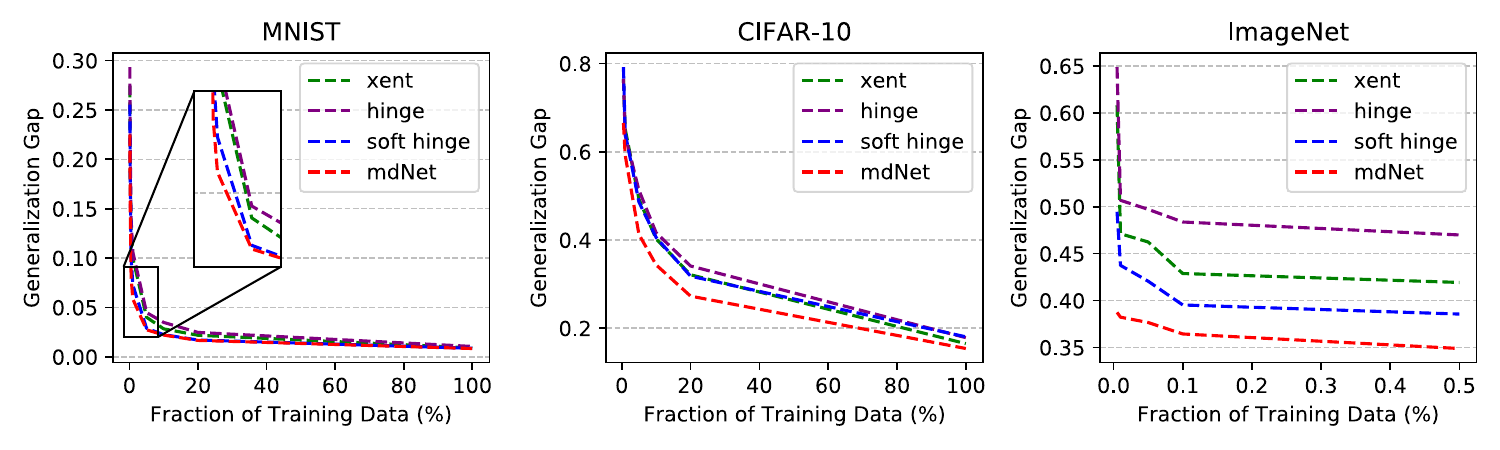}
	\caption{Compare the convergence rate of generalization gap with the increase of training samples under different loss functions on MNIST, CIFAR-10 and ImageNet datasets.}
	\label{fig:convergence}
	\vskip 0.0in
\end{figure}

Therefore, we design experiments to compare the empirical value of generalization gap with the increase of training samples under different loss functions. In Figure~\ref{fig:convergence}, it shows that the red dotted line representing the convergence curve of our method always convergences faster than the other lines under different datasets and models. It also demonstrates that our method can effectively control the capacity of the model by optimizing the margin distribution ratio, so that the trained model has better generalization performance.

Given a fixed number of samples $m$, we find that the worst case of generalization gap is proportional to the model capacity. When $m$ is large enough, the scale factor $1/\sqrt{m}$ will be close to 0, and the difference of sample complexity is not significant. The advantage of optimizing margin distribution is relatively significant when $\sqrt{m}$ is relatively small. Therefore, in the right of Figure~\ref{fig:convergence} (the ImageNet experiment), we specially truncate the most significant result of convergence rate (form 0.1\textperthousand \ to 5\textperthousand \ of training set), which shows that optimizing margin distribution can control the capacity of the model even on such a complex dataset.

\subsection{Influence of the hyper-parameters}\label{sec:hyper}

\begin{figure*}[!h]
	\centering
	\subfigure[LeNet for MNIST.]{\label{fig:hyper1}
		\begin{minipage}[]{0.30\textwidth}
			\centering
			\includegraphics[width=\textwidth]{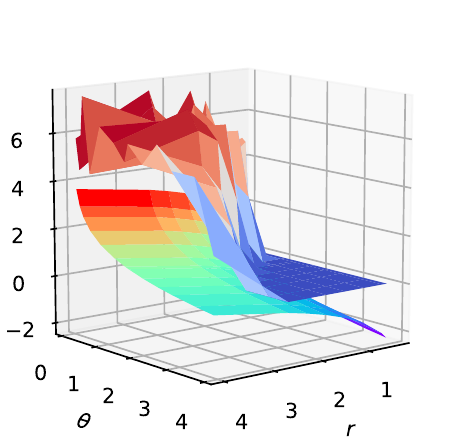}
	\end{minipage}}
	\hfill
	\subfigure[AlexNet for CIFAR-10.]{\label{fig:hyper2}
		\begin{minipage}[]{0.30\textwidth}
			\centering
			\includegraphics[width=\textwidth]{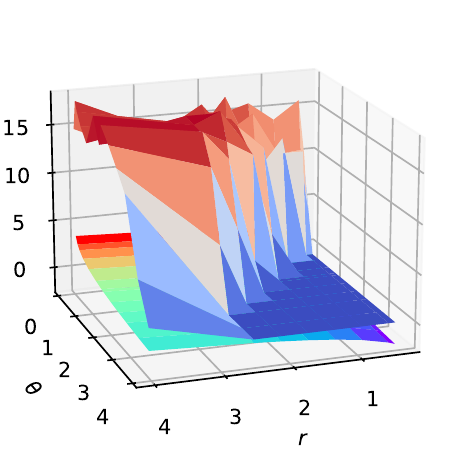}
	\end{minipage}}
	\hfill
	\subfigure[ResNet-18 for ImageNet.]{\label{fig:hyper3}
		\begin{minipage}[]{0.30\textwidth}
			\centering
			\includegraphics[width=\textwidth]{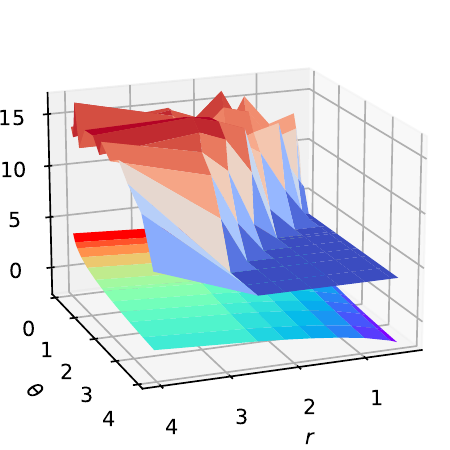}
	\end{minipage}}
	\caption{The test accuracy varying with two hyper-parameter $r$ and $\theta$ on MNIST, CIFAR-10 and ImageNet datasets. The logarithm of ratio $\ln (1/\lambda)=\ln (r/\theta)$ is the lower surface with rainbow colors and the test accuracy is the upper surface with warm-cool colors. The test accuracy is rescaled to be within the same rage as the ratio.}
	\label{fig:hyper}
	\vskip -0.0in
\end{figure*}

Figure~\ref{fig:hyper} plots the 3D surface figure for the test accuracy on the MNIST, CIFAR-10 and ImageNet datasets varying with two hyper-parameter $r$ and $\theta$. It shows that the ratio $1/\lambda=r/\theta$ (the lower surface with rainbow colors) increases with $r$ increasing and $\theta$ decreasing. As for the test accuracy (the upper surface with warm-cool colors), we find that its trend is consistent with the ratio $1/\lambda$. Therefore, the influence of the hyper-parameters demonstrates that our theoretical result. Within a certain range, getting a smaller ratio $\lambda$ through specific optimization (the margin distribution loss function) will effectively reduce the size of the hypothesis set for deep nets (returned by the specific algorithm), so as to improve the generalization ability of the learned model. In other words, the test accuracy changes consistently with the ratio of hyper-parameters (an estimation of the ratio of margin distribution). The parameter $\eta$ to trade off two different kinds of deviation (keep the balance on both sides of the margin mean) is always fixed to $0.1$ in practice.

\section{Conclusion} \label{sec:cc}

This paper proves generalization bound for deep neural networks by considering the margin distribution at the last layer instead of the minimum margin. The theoretical result inspires us to utilize a margin distribution loss function to improve the generalization performance of neural networks. Experimental results show that our method can effectively control the model capacity by optimizing the margin distribution measure, so that the trained model learns more separable representations and has better generalization performance. In future work, we will explore the effectiveness of regularization methods from a \textit{margin theory} perspective.

\section*{Acknowledgement}
This research was supported by the National Science Foundation of China (61921006) and the Collaborative Innovation Center of Novel Software Technology and Industrialization. The authors would like to thank the anonymous reviewers for constructive suggestions, as well as Yi-Xiao He and Yi-He Chen for helpful discussions.

\begin{table}[b]
	\centering
	\caption{Test accuracy of LeNet on MNIST, AlexNet on CIFAR-10 and ResNet-18 on ImageNet datasets with different regularization methods and different fractions of training set. The best accuracy on each training dataset is highlighted in bold red type. The bold black text indicates the better accuracy between the two losses with the same regularization.}
	\resizebox{0.8\textwidth}{!}{
		\begin{tabular}{ccccc}
			\toprule
			\multicolumn{5}{c}{MNSIT} \\
			\midrule
			\multicolumn{1}{c}{\multirow{2}[3]{*}{Accuracy(\%)}} & xent  & hinge & soft hinge & mdNet \\
			\cmidrule{2-5}    \multicolumn{1}{c}{} & \multicolumn{4}{c}{Batch Normalization} \\ \midrule
			100\%\_Dropout & 99.095 ± 0.083 & 98.593 ± 0.164 & 99.148 ± 0.039 & \textcolor[rgb]{ 1,  0,  0}{\textbf{99.161 ± 0.073}} \\
			100\%\_Non\_Dropout & 98.384 ± 0.072 & 97.571 ± 0.178 & 98.475 ± 0.064 & \textbf{98.837 ± 0.091} \\
			\midrule
			5\%\_Dropout & 97.001 ± 0.131 & 96.527 ± 0.219 & 97.112 ± 0.092 & \textcolor[rgb]{ 1,  0,  0}{\textbf{97.268 ± 0.113}} \\
			5\%\_Non\_Dropout & 83.364 ± 0.452 & 83.292 ± 0.721 & 83.749 ± 0.273 & \textbf{84.483 ± 0.348} \\
			\midrule
			& \multicolumn{4}{c}{Non Batch Normalization} \\ \midrule
			100\%\_Dropout & 98.228 ± 0.079 & 98.029 ± 0.184 & 98.271 ± 0.055 & \textbf{98.342 ± 0.069} \\
			100\%\_Non\_Dropout & 91.728 ± 0.117 & 90.237 ± 0.318 & 91.029 ± 0.098 & \textbf{92.274 ± 0.121} \\
			\midrule
			5\%\_Dropout & 77.842 ± 0.489 & 76.938 ± 0.827 & 77.727 ± 0.411 & \textbf{78.173 ± 0.619} \\
			5\%\_Non\_Dropout & 58.023 ± 0.951 & 57.822 ± 1.280 & 59.384 ± 0.827 & \textbf{61.379 ± 0.588} \\
			\midrule
			\multicolumn{5}{c}{CIFAR-10} \\
			\midrule
			\multicolumn{1}{c}{\multirow{2}[4]{*}{Accuracy(\%)}} & xent  & hinge & soft hinge & mdNet \\
			\cmidrule{2-5}    \multicolumn{1}{c}{} & \multicolumn{4}{c}{Batch Normalization} \\
			\midrule
			100\%\_Dropout & 85.782 ± 0.198 & 84.234 ± 0.748 & 86.744 ± 0.294 & \textcolor[rgb]{ 1,  0,  0}{\textbf{87.644 ± 0.151}} \\
			100\%\_Non\_Dropout & 81.491 ± 0.143 & 80.938 ± 0.812 & 86.032 ± 0.298 & \textbf{86.233 ± 0.244} \\
			\midrule
			5\%\_Dropout & 61.955 ± 1.945 & 58.363 ± 2.450 & 59.441 ± 1.316 & \textcolor[rgb]{ 1,  0,  0}{\textbf{67.636 ± 1.633}} \\
			5\%\_Non\_Dropout & 57.753 ± 2.228 & 54.289 ± 3.482 & 56.839 ± 2.318 & \textbf{64.173 ± 1.982} \\
			\midrule
			& \multicolumn{4}{c}{Non Batch Normalization} \\
			\midrule
			100\%\_Dropout & 83.517 ± 0.322 & 82.153 ± 1.236 & 81.961 ± 0.293 & \textbf{84.643 ± 0.255} \\
			100\%\_Non\_Dropout & 72.223 ± 1.284 & 69.379 ± 2.907 & 75.267 ± 1.027 & \textbf{76.793 ± 1.279} \\
			\midrule
			5\%\_Dropout & 50.747 ± 3.735 & 42.739 ± 6.763 & 52.847 ± 1.823 & \textbf{58.739 ± 1.348} \\
			5\%\_Non\_Dropout & 36.293 ± 4.872 & 30.984 ± 7.736 & 43.265 ± 4.263 & \textbf{47.056 ± 3.927} \\
			\midrule
			\multicolumn{5}{c}{ImageNet} \\
			\midrule
			& xent         & hinge        & soft hinge   & mdNet                                        \\ \cmidrule(l){2-5}  
			\multirow{-2}{*}{Accuracy(\%)} & \multicolumn{4}{c}{Batch Normalization}                                                   \\ \midrule
			100\%\_Dropout                 & 70.238 ± 1.221 & 69.782 ± 1.933  & 70.284 ± 1.022 & {\color[HTML]{FF0000} \textbf{70.758 ± 1.014}} \\
			100\%\_Non\_Dropout            & 68.484 ± 1.265 & 67.918 ± 2.166 & 68.83 ± 1.151 & \textbf{69.447 ± 1.124}                        \\ \midrule
			5‰\_Dropout                    & 60.176 ± 2.045 & 57.475 ± 2.023 & 61.379 ± 1.053 & {\color[HTML]{FF0000} \textbf{65.080 ± 2.373}} \\
			5‰\_Non\_Dropout               & 59.574 ± 2.747 & 56.621 ± 2.253 & 60.068 ± 1.773 & \textbf{63.529 ± 2.012}                        \\ \midrule
			& \multicolumn{4}{c}{Non Batch Normalization}                                               \\ \midrule
			100\%\_Dropout                 & 66.924 ± 1.552 & 67.387 ± 1.764 & 67.462 ± 1.017 & \textbf{68.655 ± 1.732}                        \\
			100\%\_Non\_Dropout            & 64.481 ± 2.183 & 61.820 ± 2.947 & 64.334 ± 2.367 & \textbf{65.838 ± 2.481}                        \\ \midrule
			5‰\_Dropout                    & 54.961 ± 3.382 & 52.543 ± 3.722 & 55.757 ± 2.357 & \textbf{58.774 ± 3.841}                        \\
			5‰\_Non\_Dropout               & 47.374 ± 3.265 & 45.741 ± 5.349 & 48.798 ± 3.392 & \textbf{53.727 ± 4.235}  \\\bottomrule
		\end{tabular}
	}%
	\label{tab:ablation}
\end{table}


\nocite{*}
\printbibliography

@inproceedings{gronlund2019margin,
  author    = {Allan Gr{\o}nlund and
               Lior Kamma and
               Kasper Green Larsen and
               Alexander Mathiasen and
               Jelani Nelson},
  title     = {Margin-Based Generalization Lower Bounds for Boosted Classifiers},
  booktitle = {Advances in Neural Information Processing Systems 32},
  pages     = {11940--11949},
  year      = {2019}
}

@inproceedings{zhou2014large,
  author    = {Zhi{-}Hua Zhou},
  title     = {Large Margin Distribution Learning},
  booktitle = {Artificial Neural Networks in Pattern Recognition},
  pages     = {1--11},
  year      = {2014}
}

@ARTICLE{gori2022ten,
AUTHOR={Gori, Marco},    
TITLE={Ten Questions for a Theory of Vision},      
JOURNAL={Frontiers in Computer Science},      
VOLUME={3}, 
pages = {701248},     
YEAR={2022},
DOI={10.3389/fcomp.2021.701248}   
}

@article{wu2021learning,
  author    = {Wei Wu and
               Xiaoyuan Jing and
               Wencai Du and
               Guoliang Chen},
  title     = {Learning dynamics of gradient descent optimization in deep neural
               networks},
  journal   = {Science China Information Science},
  volume    = {64},
  number    = {5},
  year      = {2021},
  pages     = {15102},
  doi      = {10.1007/s11432-020-3163-0}
}

@ARTICLE{matskevych2022from,
  
AUTHOR={Matskevych, Alex and Wolny, Adrian and Pape, Constantin and Kreshuk, Anna},   
   
TITLE={From Shallow to Deep: {E}xploiting Feature-Based Classifiers for Domain Adaptation in Semantic Segmentation},      
  
JOURNAL={Frontiers in Computer Science},      
  
VOLUME={4},      
pages = {805166},
  
YEAR={2022},

doi = {10.3389/fcomp.2022.805166}
}

@article{zhou2021over,
  title={Why over-parameterization of deep neural networks does not overfit?},
  author={Zhou, Zhi-Hua},
  journal={Science China Information Sciences},
  volume={64},
  number={1},
  pages={1--3},
  year={2021}
}

@book{glantz2001primer,
  title={Primer of Applied Regression \& Analysis of Variance, ed},
  author={Glantz, Stanton and Slinker, Bryan},
  year={2001},
  publisher={McGraw-Hill, Inc., New York}
}

@book{mohri2018foundations,
  title={Foundations of Machine Learning},
  author={Mohri, Mehryar and Rostamizadeh, Afshin and Talwalkar, Ameet},
  year={2018},
  publisher={MIT press}
}

@article{zhang2012generalization,
  title={Generalization bounds for domain adaptation},
  author={Zhang, Chao and Zhang, Lei and Ye, Jieping},
  journal={Advances in Neural Information Processing Systems 25},
  volume={4},
  pages={3320},
  year={2012}
}

@inproceedings{mansour2009domain,
  author    = {Yishay Mansour and
               Mehryar Mohri and
               Afshin Rostamizadeh},
  title     = {Domain Adaptation: Learning Bounds and Algorithms},
  booktitle = {The 22nd Conference on Learning Theory},
  year      = {2009}
}

@article{mansour2014robust,
  author    = {Yishay Mansour and
               Mariano Schain},
  title     = {Robust domain adaptation},
  journal   = {Annals of Mathematics and Artificial Intelligence},
  volume    = {71},
  number    = {4},
  pages     = {365--380},
  year      = {2014}
}

@inproceedings{becker2013nonlinear,
  author    = {Carlos J. Becker and
               C. Mario Christoudias and
               Pascal Fua},
  title     = {Non-Linear Domain Adaptation with Boosting},
  booktitle = {Advances in Neural Information Processing Systems 26},
  pages     = {485--493},
  year      = {2013}
}

@article{rozantsev2019beyond,
  author    = {Artem Rozantsev and
               Mathieu Salzmann and
               Pascal Fua},
  title     = {Beyond Sharing Weights for Deep Domain Adaptation},
  journal   = {{IEEE} Transactions on Pattern Analysis and Machine Intelligence},
  volume    = {41},
  number    = {4},
  pages     = {801--814},
  year      = {2019}
}

@inproceedings{koniusz2017domain,
  author    = {Piotr Koniusz and
               Yusuf Tas and
               Fatih Porikli},
  title     = {Domain Adaptation by Mixture of Alignments of Second-or Higher-Order
               Scatter Tensors},
  booktitle = {{IEEE} Conference on Computer Vision and Pattern Recognition},
  pages     = {7139--7148},
  year      = {2017}
}

@inproceedings{pan2009domain,
  author    = {Sinno Jialin Pan and
               Ivor W. Tsang and
               James T. Kwok and
               Qiang Yang},
  title     = {Domain Adaptation via Transfer Component Analysis},
  booktitle = {Proceedings of the 21st International Joint Conference
               on Artificial Intelligence},
  pages     = {1187--1192},
  year      = {2009}
}

@inproceedings{wang2021generalization,
  author    = {Jindong Wang and
               Cuiling Lan and
               Chang Liu and
               Yidong Ouyang and
               Tao Qin},
  title     = {Generalizing to Unseen Domains: {A} Survey on Domain Generalization},
  booktitle = {Proceedings of the 30th International Joint Conference on Artificial
               Intelligence},
  pages     = {4627--4635},
  year      = {2021}
}

@inproceedings{muandet2013domain,
  author    = {Krikamol Muandet and
               David Balduzzi and
               Bernhard Sch{\"{o}}lkopf},
  title     = {Domain Generalization via Invariant Feature Representation},
  booktitle = {Proceedings of the 30th International Conference on Machine Learning
              },
  volume    = {28},
  pages     = {10--18},
  publisher = {JMLR.org},
  year      = {2013}
}

@inproceedings{ghifary2015domain,
  author    = {Muhammad Ghifary and
               W. Bastiaan Kleijn and
               Mengjie Zhang and
               David Balduzzi},
  title     = {Domain Generalization for Object Recognition with Multi-task Autoencoders},
  booktitle = {{IEEE} International Conference on Computer Vision},
  pages     = {2551--2559},
  year      = {2015}
}

@inproceedings{dubey2021adaptive,
  author    = {Abhimanyu Dubey and
               Vignesh Ramanathan and
               Alex Pentland and
               Dhruv Mahajan},
  title     = {Adaptive Methods for Real-World Domain Generalization},
  booktitle = {{IEEE} Conference on Computer Vision and Pattern Recognition},
  pages     = {14340--14349},
  year      = {2021}
}

@inproceedings{NeyshaburBMS17,
  author    = {Behnam Neyshabur and
               Srinadh Bhojanapalli and
               David McAllester and
               Nati Srebro},
  title     = {Exploring Generalization in Deep Learning},
  booktitle = {Advances in Neural Information Processing Systems 30,
              },
  pages     = {5947--5956},
  year      = {2017}
}

@inproceedings{ZhuWYWM19,
  author    = {Zhanxing Zhu and
               Jingfeng Wu and
               Bing Yu and
               Lei Wu and
               Jinwen Ma},
  title     = {The Anisotropic Noise in Stochastic Gradient Descent: Its Behavior
               of Escaping from Sharp Minima and Regularization Effects},
  booktitle = {Proceedings of the 36th International Conference on Machine Learning},
  volume    = {97},
  pages     = {7654--7663},
  year      = {2019}
}

@inproceedings{DinhPBB17,
  author    = {Laurent Dinh and
               Razvan Pascanu and
               Samy Bengio and
               Yoshua Bengio},
  title     = {Sharp Minima Can Generalize For Deep Nets},
  booktitle = {Proceedings of the 34th International Conference on Machine Learning},
  volume    = {70},
  pages     = {1019--1028},
  year      = {2017}
}

@inproceedings{KeskarMNST17,
  author    = {Nitish Shirish Keskar and
               Dheevatsa Mudigere and
               Jorge Nocedal and
               Mikhail Smelyanskiy and
               Ping Tak Peter Tang},
  title     = {On Large-Batch Training for Deep Learning: Generalization Gap and
               Sharp Minima},
  booktitle = {5th International Conference on Learning Representations},
  year      = {2017}
}

@article{chen2018stability,
  title={Stability and convergence trade-off of iterative optimization algorithms},
  author={Chen, Yuansi and Jin, Chi and Yu, Bin},
  journal   = {CoRR},  
  volume    = {abs/1804.01619},
  year={2018}
}

@inproceedings{Mou0Z018,
  author    = {Wenlong Mou and
               Liwei Wang and
               Xiyu Zhai and
               Kai Zheng},
  title     = {Generalization Bounds of {SGLD} for Non-convex Learning: Two Theoretical
               Viewpoints},
  booktitle = {Conference On Learning Theory},
  volume    = {75},
  pages     = {605--638},
  year      = {2018}
}

@inproceedings{HardtRS16,
  author    = {Moritz Hardt and
               Ben Recht and
               Yoram Singer},
  title     = {Train faster, generalize better: Stability of stochastic gradient
               descent},
  booktitle = {Proceedings of the 33nd International Conference on Machine Learning},
  volume    = {48},
  pages     = {1225--1234},
  year      = {2016}
}

@article{wei2018margin,
  author    = {Colin Wei and
               Jason D. Lee and
               Qiang Liu and
               Tengyu Ma},
  title     = {On the Margin Theory of Feedforward Neural Networks},
  journal   = {CoRR},
  volume    = {abs/1810.05369},
  year      = {2018}
}

@inproceedings{li2018algorithmic,
  author    = {Yuanzhi Li and
               Tengyu Ma and
               Hongyang Zhang},
  title     = {Algorithmic Regularization in Over-parameterized Matrix Sensing and
               Neural Networks with Quadratic Activations},
  booktitle = {Conference On Learning Theory},
  volume    = {75},
  pages     = {2--47},
  year      = {2018}
}

@inproceedings{ji2019gradient,
  author    = {Ziwei Ji and
               Matus Telgarsky},
  title     = {Gradient descent aligns the layers of deep linear networks},
  booktitle = {7th International Conference on Learning Representations},
  year      = {2019}
}

@inproceedings{gunasekar2018characterizing,
  author    = {Suriya Gunasekar and
               Jason D. Lee and
               Daniel Soudry and
               Nathan Srebro},
  title     = {Characterizing Implicit Bias in Terms of Optimization Geometry},
  booktitle = {Proceedings of the 35th International Conference on Machine Learning
               },
  volume    = {80},
  pages     = {1827--1836},
  year      = {2018},
}

@inproceedings{gunasekar2018implicit,
  author    = {Suriya Gunasekar and
               Jason D. Lee and
               Daniel Soudry and
               Nati Srebro},
  title     = {Implicit Bias of Gradient Descent on Linear Convolutional Networks},
  booktitle = {Advances in Neural Information Processing Systems 31},
  pages     = {9482--9491},
  year      = {2018}
}

@article{soudry2018implicit,
  author    = {Daniel Soudry and
               Elad Hoffer and
               Mor Shpigel Nacson and
               Suriya Gunasekar and
               Nathan Srebro},
  title     = {The Implicit Bias of Gradient Descent on Separable Data},
  journal   = {Journal of Machine Learning Research},
  volume    = {19},
    number  = {70},
  pages     = {1--57},
  year      = {2018}
}

@article{bartlett1998almost,
  title={Almost linear {VC}-dimension bounds for piecewise polynomial networks},
  author={Bartlett, Peter L and Maiorov, Vitaly and Meir, Ron},
  journal={Neural Computation},
  volume={10},
  number={8},
  pages={2159--2173},
  year={1998}
}

@inproceedings{zhang2020optimal,
  author    = {Teng Zhang and
               Peng Zhao and
               Hai Jin},
  title     = {Optimal Margin Distribution Learning in Dynamic Environments},
  booktitle = {Proceedings of the 34th {AAAI} Conference on Artificial Intelligence},
  pages     = {6821--6828},
  year      = {2020}
}

@inproceedings{zhang18optimal,
  author    = {Teng Zhang and
               Zhi{-}Hua Zhou},
  title     = {Optimal Margin Distribution Clustering},
  booktitle = {Proceedings of the 32nd {AAAI} Conference on Artificial Intelligence},
  pages     = {4474--4481},
  year      = {2018}
}

@inproceedings{zhang18semi,
  author    = {Teng Zhang and
               Zhi{-}Hua Zhou},
  title     = {Semi-Supervised Optimal Margin Distribution Machines},
  booktitle = {Proceedings of the 27th International Joint Conference on Artificial Intelligence},
  pages     = {3104--3110},
  year      = {2018}
}

@article{tan2020multi,
  title={Multi-label optimal margin distribution machine},
  author={Tan, Zhi-Hao and Tan, Peng and Jiang, Yuan and Zhou, Zhi-Hua},
  journal={Machine Learning},
  volume={109},
  number={3},
  pages={623--642},
  year={2020}
}

@inproceedings{goodfellow2015explaining,
  author    = {Ian J. Goodfellow and
               Jonathon Shlens and
               Christian Szegedy},
  title     = {Explaining and Harnessing Adversarial Examples},
  booktitle = {3rd International Conference on Learning Representations},
  year      = {2015}
}

@inproceedings{papernot2017practical,
  author    = {Nicolas Papernot and
               Patrick D. McDaniel and
               Ian J. Goodfellow and
               Somesh Jha and
               Z. Berkay Celik and
               Ananthram Swami},
  title     = {Practical Black-Box Attacks against Machine Learning},
  booktitle = {Proceedings of the 2017 {ACM} on Asia Conference on Computer and Communications
               Security},
  pages     = {506--519},
  year      = {2017}
}

@article{zhang2021understanding,
  author    = {Chiyuan Zhang and
               Samy Bengio and
               Moritz Hardt and
               Benjamin Recht and
               Oriol Vinyals},
  title     = {Understanding deep learning (still) requires rethinking generalization},
  journal   = {Communications of the ACM},
  volume    = {64},
  number    = {3},
  pages     = {107--115},
  year      = {2021}
}

@article{azulay2019why,
  author    = {Aharon Azulay and
               Yair Weiss},
  title     = {Why do deep convolutional networks generalize so poorly to small image
               transformations?},
  journal   = {Journal of Machine Learning Research},
  year    = {2019},
  volume  = {20},
  number  = {184},
  pages   = {1-25},
}

@article{deng2012imagenet,
Author = {Olga Russakovsky and Jia Deng and Hao Su and Jonathan Krause and Sanjeev Satheesh and Sean Ma and Zhiheng Huang and Andrej Karpathy and Aditya Khosla and Michael Bernstein and Alexander C. Berg and Li Fei-Fei},
Title = {ImageNet Large Scale Visual Recognition Challenge},
Year = {2015},
journal   = {International Journal of Computer Vision},
volume={115},
number={3},
pages={211-252}
}

@inproceedings{ustc,
  title     = {Theoretical Investigation of Generalization Bound for Residual Networks},
  author    = {Chen, Hao and Mo, Zhanfeng and Yang, Zhouwang and Wang, Xiao},
  booktitle = {Proceedings of the 28th International Joint Conference on
               Artificial Intelligence},      
  pages     = {2081--2087},
  year      = {2019}
}

@inproceedings{paszke2017automatic,
  title = {PyTorch: An Imperative Style, High-Performance Deep Learning Library},
author = {Paszke, Adam and Gross, Sam and Massa, Francisco and Lerer, Adam and Bradbury, James and Chanan, Gregory and Killeen, Trevor and Lin, Zeming and Gimelshein, Natalia and Antiga, Luca and Desmaison, Alban and Kopf, Andreas and Yang, Edward and DeVito, Zachary and Raison, Martin and Tejani, Alykhan and Chilamkurthy, Sasank and Steiner, Benoit and Fang, Lu and Bai, Junjie and Chintala, Soumith},
booktitle = {Advances in Neural Information Processing Systems 32},
pages = {8024--8035},
year = {2019}
}

@article{bousquet02stability,
  title={Stability and generalization},
  author={Bousquet, Olivier and Elisseeff, Andr{\'e}},
  journal={Journal of Machine Learning Research},
  volume={2},
  number={Mar},
  pages={499--526},
  year={2002}
}

@inproceedings{neyshabur15norm,
  title={Norm-based capacity control in neural networks},
  author={Neyshabur, Behnam and Tomioka, Ryota and Srebro, Nathan},
  booktitle={Proceedings of the 28th Annual Conference on Learning Theory},
  pages={1376--1401},
  year={2015}
}

@article{bartlett02rademacher,
  title={Rademacher and {G}aussian complexities: Risk bounds and structural results},
  author={Bartlett, Peter L and Mendelson, Shahar},
  journal={Journal of Machine Learning Research},
  volume={3},
  pages={463--482},
  year={2002}
}

@article{lecun15deep,
  title={Deep learning},
  author={LeCun, Yann and Bengio, Yoshua and Hinton, Geoffrey},
  journal={Nature},
  volume={521},
  number={7553},
  pages={436},
  year={2015}
}

@InProceedings{harvey17nearly,
  title =    {Nearly-tight {VC}-dimension bounds for piecewise linear neural networks},
  author =   {Nick Harvey and Christopher Liaw and Abbas Mehrabian},
  booktitle =    {Proceedings of the 30th Annual Conference on Learning Theory},
  pages =    {1064--1068},
  year =   {2017}
}

@book{mohri18foundations,
  title={Foundations of Machine Learning},
  author={Mohri, Mehryar and Rostamizadeh, Afshin and Talwalkar, Ameet},
  year={2018},
  publisher={MIT press}
}

@inproceedings{baldi13understanding,
  title={Understanding dropout},
  author={Baldi, Pierre and Sadowski, Peter J},
  booktitle={Advances in Neural Information Processing Systems 27},
  pages={2814--2822},
  year={2013}
}

@inproceedings{he16res,
  author    = {Kaiming He and
               Xiangyu Zhang and
               Shaoqing Ren and
               Jian Sun},
  title     = {Deep Residual Learning for Image Recognition},
  booktitle = {Proceedings of the IEEE Computer Society Conference on Computer Vision and Pattern Recognition},
  pages     = {770--778},
  year      = {2016}
}

@inproceedings{karen15very,
title={Very Deep Convolutional Networks for Large-Scale Image Recognition},
author={Karen Simonyan and Andrew Zisserman},
booktitle={3rd International Conference on Learning Representations},
year={2015}
}

@InProceedings{huang18learning,
  title = 	 {Learning Deep {R}es{N}et Blocks Sequentially using Boosting Theory},
  author = 	 {Huang, Furong and Ash, Jordan and Langford, John and Schapire, Robert},
  booktitle = 	 {Proceedings of the 35th International Conference on Machine Learning},
  pages = 	 {2058--2067},
  year = 	 {2018}
}

@inproceedings{jiang19predicting,
title={Predicting the Generalization Gap in Deep Networks with Margin Distributions},
author={Yiding Jiang and Dilip Krishnan and Hossein Mobahi and Samy Bengio},
booktitle={7th International Conference on Learning Representations},
year={2019}
}

@inproceedings{liu16large,
  title={Large-Margin Softmax Loss for Convolutional Neural Networks},
  author={Liu, Weiyang and Wen, Yandong and Yu, Zhiding and Yang, Meng},
  booktitle={Proceedings of the 33rd International Conference on Machine Learning},
  pages={507--516},
  year={2016}
}

@inproceedings{sun14deep,
  title={Deep learning face representation by joint identification-verification},
  author={Sun, Yi and Chen, Yuheng and Wang, Xiaogang and Tang, Xiaoou},
  booktitle={Advances in Neural Information Processing Systems 28},
  pages={1988--1996},
  year={2014}
}

@inproceedings{schroff15facenet,
  title={Facenet: {A} unified embedding for face recognition and clustering},
  author={Schroff, Florian and Kalenichenko, Dmitry and Philbin, James},
  booktitle={Proceedings of the IEEE Computer Society Conference on Computer Vision and Pattern Recognition},
  pages={815--823},
  year={2015}
}

@article{chan15pcanet,
  title={P{CAN}et: {A} simple deep learning baseline for image classification?},
  author={Chan, Tsung-Han and Jia, Kui and Gao, Shenghua and Lu, Jiwen and Zeng, Zinan and Ma, Yi},
  journal={IEEE Transactions on Image Processing},
  volume={24},
  number={12},
  pages={5017--5032},
  year={2015}
}

@article{maaten08visualizing,
  title={Visualizing data using t-{SNE}},
  author={van der Maaten, Laurens and Hinton, Geoffrey},
  journal={Journal of Machine Learning Research},
  volume={9},
  number={11},
  pages={2579--2605},
  year={2008}
}

@techreport{krizhevsky09learning,
  title={Learning Multiple Layers of Features from Tiny Images},
  author={Alex Krizhevsky},
  year={2009}
}

@article{lecun98gradient,
  title={Gradient-based learning applied to document recognition},
  author={LeCun, Yann and Bottou, L{\'e}on and Bengio, Yoshua and Haffner, Patrick},
  journal={Proceedings of the IEEE},
  booktitle = {Proceedings of the IEEE},
  volume={86},
  number={11},
  pages={2278--2324},
  year={1998},
}

@incollection{alex12imagenet,
	title = {ImageNet Classification with Deep Convolutional Neural Networks},
	author = {Alex Krizhevsky and Sutskever, Ilya and Hinton, Geoffrey E},
	booktitle = {Advances in Neural Information Processing Systems 26},
	pages = {1097--1105},
	year = {2012},
}

@inproceedings{krogh92simple,
  title={A simple weight decay can improve generalization},
  author={Krogh, Anders and Hertz, John A},
  booktitle={Advances in Neural Information Processing Systems 5},
  pages={950--957},
  year={1992}
}

@inproceedings{ioffe15batch,
  author    = {Sergey Ioffe and
               Christian Szegedy},
  title     = {Batch Normalization: Accelerating Deep Network Training by Reducing
               Internal Covariate Shift},
  booktitle = {Proceedings of the 32nd International Conference on Machine Learning},
  pages     = {448--456},
  year      = {2015},
}

@article{srivastava14dropout,
  title={Dropout: {a} simple way to prevent neural networks from overfitting},
  author={Srivastava, Nitish and Hinton, Geoffrey and Krizhevsky, Alex and Sutskever, Ilya and Salakhutdinov, Ruslan},
  journal={Journal of Machine Learning Research},
  volume={15},
  number={1},
  pages={1929--1958},
  year={2014}
}

@aticle{wang11boosting,
	 author = {Wang, Liwei and Sugiyama, Masashi and Jing, Zhaoxiang and Yang, Cheng and Zhou, Zhi-Hua and Feng, Jufu},
	 title = {A Refined Margin Analysis for Boosting Algorithms via Equilibrium Margin},
	 journal = {Journal of Machine Learning Research},
	 volume = {12},
	 year = {2011},
	 pages = {1835--1863},
	 numpages = {29},
}

@inproceedings{reyzin06boosting,
	 author = {Reyzin, Lev and Schapire, Robert E.},
	 title = {How Boosting the Margin Can Also Boost Classifier Complexity},
	 booktitle = {Proceedings of the 23rd International Conference on Machine Learning},
	 year = {2006},
	 pages = {753--760},
	 numpages = {8}
}

@article{breiman99arc,
	 author    = {Leo Breiman},
	  title     = {Prediction Games and {A}rcing Algorithms},
	  journal   = {Neural Computation},
	  booktitle = {Neural Computation},
	  volume    = {11},
	  number    = {7},
	  pages     = {1493--1517},
	 year = {1999}
}

@inproceedings{schapire97boosting,
  author    = {Robert E. Schapire and
               Yoav Freund and
               Peter Barlett and
               Wee Sun Lee},
  title     = {Boosting the margin: {A} new explanation for the effectiveness of
               voting methods},
  booktitle = {Proceedings of the 14th International Conference on Machine
               Learning},
  pages     = {322--330},
  year      = {1997},
}

@inproceedings{bartlett17spectrally,
  author    = {Peter L. Bartlett and
               Dylan J. Foster and
               Matus J. Telgarsky},
  title     = {Spectrally-normalized margin bounds for neural networks},
  booktitle = {Advances in Neural Information Processing Systems 31},
  pages     = {6241--6250},
  year      = {2017}
}

@article{cortes95svm,
	author="Cortes, Corinna
	and Vapnik, Vladimir",
	title="Support-Vector Networks",
	booktitle = {Machine Learning},
	journal="Machine Learning",
	year="1995",
	volume="20",
	number="3",
	pages="273--297"
}

@book{vapnik98statistical,
  author    = {Vladimir Vapnik},
  title     = {Statistical Learning Theory},
  publisher = {Wiley press},
  year      = {1998}
}

@inproceedings{harris97svrg,
	title = {Support Vector Regression Machines},
	author = {Harris Drucker and Burges, Christopher J. C. and Linda Kaufman and Alex J. Smola and Vladimir Vapnik},
	booktitle = {Advances in Neural Information Processing Systems 10},
	pages = {155--161},
	year = {1997}
}

@inproceedings{arora18compression,
  author    = {Sanjeev Arora and
               Rong Ge and
               Behnam Neyshabur and
               Yi Zhang},
  title     = {Stronger Generalization Bounds for Deep Nets via a Compression Approach},
  booktitle = {Proceedings of the 35th International Conference on Machine Learning},
  pages     = {254--263},
  year      = {2018}
}

@inproceedings{neyshabur18spectrally,
title={A {PAC}-{B}ayesian Approach to Spectrally-Normalized Margin Bounds for Neural Networks},
author={Behnam Neyshabur and Srinadh Bhojanapalli and Nathan Srebro},
booktitle={6th International Conference on Learning Representations},
year={2018}
}

@inproceedings{mcalleter03margin,
	author="McAllester, David",
	title="Simplified {PAC}-{B}ayesian Margin Bounds",
	booktitle="Learning Theory and Kernel Machines",
	year="2003",
	pages="203--215",
}

@inproceedings{elsayed18large,
title = {Large Margin Deep Networks for Classification},
author = {Elsayed, Gamaleldin and Krishnan, Dilip and Mobahi, Hossein and Regan, Kevin and Bengio, Samy},
booktitle = {Advances in Neural Information Processing Systems 31},
pages = {850--860},
year = {2018}
}

@article{zhang16odm,
  title={Optimal margin distribution machine},
  author={Zhang, Teng and Zhou, Zhi-Hua},
  journal={IEEE Transactions on Knowledge and Data Engineering},
  volume={32},
  number={6},
  pages={1143--1156},
  year={2019}
}

@article{gao13boosting,
	 author = {Gao, Wei and Zhou, Zhi-Hua},
	 title = {On the Doubt About Margin Explanation of Boosting},
	 booktitle = {Artificial Intelligence},
	 journal = {Artificial Intelligence},
	 year = {2013},
	 volume = {203},
	 pages = {1--18},
	 numpages = {18}
}

@inproceedings{zhang17mcodm,
  title = 	 {Multi-Class Optimal Margin Distribution Machine},
  author = 	 {Teng Zhang and Zhi-Hua Zhou},
  booktitle = 	 {Proceedings of the 34th International Conference on Machine Learning},
  pages = 	 {4063--4071},
  year = 	 {2017}
}

@article{mackey14matrix,
  title={Matrix concentration inequalities via the method of exchangeable pairs},
  author={Mackey, Lester and Jordan, Michael I and Chen, Richard Y and Farrell, Brendan and Tropp, Joel A and others},
  journal={The Annals of Probability},
  volume={42},
  number={3},
  pages={906--945},
  year={2014}
}

@article{tropp12user,
  title={User-friendly tail bounds for sums of random matrices},
  author={Tropp, Joel A},
  journal={Foundations of Computational Mathematics},
  volume={12},
  number={4},
  pages={389--434},
  year={2012}
}

@article{hinton12neural,
  title={Neural networks for machine learning lecture 6a overview of mini-batch gradient descent},
  author={Hinton, Geoffrey and Srivastava, Nitish and Swersky, Kevin},
  year={2012}
}

@article{ruder16,
  author    = {Sebastian Ruder},
  title     = {An overview of gradient descent optimization algorithms},
  journal   = {CoRR},
  volume    = {abs/1609.04747},
  year      = {2016}
}

@inproceedings{kinga15method,
  author    = {Diederik P. Kingma and
               Jimmy Ba},
  title     = {Adam: {A} Method for Stochastic Optimization},
  booktitle = {3rd International Conference on Learning Representations},
  year      = {2015}
}

@inproceedings{zhang18omdc,
  author    = {Teng Zhang and
               Zhi{-}Hua Zhou},
  title     = {Optimal Margin Distribution Clustering},
  booktitle = {Proceedings of the Thirty-Second {AAAI} Conference on Artificial Intelligence},
  pages     = {4474--4481},
  year      = {2018}
}

\appendix


\end{document}